\documentclass[10pt,twocolumn,letterpaper]{article}

\usepackage{iccv}
\usepackage{times}
\usepackage{epsfig}
\usepackage{graphicx}
\usepackage{amsmath}

\usepackage{amssymb}
\usepackage{amsthm}
\usepackage{amsfonts} 
\usepackage{booktabs}
\usepackage{algorithm}
\usepackage{algorithmic}
\usepackage{xcolor}
\usepackage{subfigure}
\usepackage[inline]{enumitem}
\usepackage{pifont}
\usepackage{etoolbox}
\usepackage{array}
\usepackage{xr}
\makeatletter
\@namedef{ver@everyshi.sty}{}
\makeatother
\usepackage{tikz,subfigure}
\usetikzlibrary{decorations.pathreplacing,positioning}
\usetikzlibrary{automata, positioning, arrows}
\usepackage{xcolor}
\usepackage{nicefrac}       
\usepackage{microtype}

\newtheorem{theorem}{Theorem}[section]

\newtheorem{lemma}[theorem]{Lemma}
\theoremstyle{definition}
\newtheorem{definition}{Definition}[section]

\newcommand{\cmark}{\ding{51}}%
\newcommand{\xmark}{\ding{55}}%
\renewcommand\and{\end{tabular}\kern-\tabcolsep\  \ \kern-\tabcolsep\begin{tabular}[t]{c}}
\newcolumntype{H}{>{\setbox0=\hbox\bgroup}c<{\egroup}@{}}

\definecolor{darkblue}{HTML}{1F4E79}
\definecolor{lightblue}{HTML}{00B0F0}
\definecolor{salmon}{HTML}{FF9C6B}
\definecolor{xiaomi}{HTML}{FF670A}
\definecolor{xablue}{HTML}{00ABF5}
\definecolor{xayellow}{HTML}{FACD00}
\definecolor{xapink}{HTML}{FF6D8A}
\definecolor{xagreen}{HTML}{00DDCB}
\definecolor{xagray}{HTML}{545454}
\definecolor{xablued}{HTML}{1F77B4}
\definecolor{xayellowd}{HTML}{FF7F0E}
\definecolor{xapinkd}{HTML}{D62728}
\definecolor{xagreend}{HTML}{2CA02C}
\definecolor{xagrayd}{HTML}{615761}

\usetikzlibrary{spy}
\usetikzlibrary{shapes.geometric, arrows, backgrounds, fit}
\tikzstyle{layer} = [rectangle, minimum width=1cm, minimum height=0.5cm,text centered, draw=black, fill=lightblue]
\tikzstyle{cell} = [rectangle, minimum width=1cm, minimum height=0.5cm,text centered, draw=black, fill=salmon]
\tikzstyle{darklayer} = [rectangle, minimum width=0.5cm, minimum height=0.5cm,text centered, draw=black, fill=darkblue]
\tikzstyle{state} = [circle, inner sep=0pt, minimum width=0.5cm, minimum height=0.5cm,text centered, draw=black, fill=darkblue]
\tikzstyle{arrow} = [thick,->,>=stealth]
\tikzstyle{op} = [circle, minimum width=0.5cm, minimum height=0.5cm,text centered, draw=black]
\tikzstyle{cell_r} = [rectangle, minimum width=1cm, minimum height=0.5cm,text centered, draw=black, fill=salmon,rotate=90]
\tikzstyle{cell_r_p} = [cell_r, fill=xapink] %
\tikzstyle{cell_r_b} = [cell_r, fill=xablue]
\tikzstyle{cell_r_y} = [cell_r, fill=xayellow]
\tikzstyle{cell_r_g} = [cell_r, fill=xagreen]
\tikzstyle{e3} = [minimum width=1.5cm]
\tikzstyle{e6} = [minimum width=2.5cm]
\usetikzlibrary{backgrounds}
\pgfdeclarelayer{foreground}
\pgfsetlayers{background,main,foreground}
\tikzset{zlevel/.style={%
    execute at begin scope={\pgfonlayer{#1}},
    execute at end scope={\endpgfonlayer}
}}



\usepackage[pagebackref=true,breaklinks=true,colorlinks,bookmarks=false]{hyperref}
\usepackage{cleveref}

\title{FairNAS: Rethinking Evaluation Fairness of Weight Sharing Neural Architecture Search}

\author{
Xiangxiang Chu\thanks{This work was done when all the authors were at Xiaomi AI Lab.} \quad Bo Zhang \quad Ruijun Xu \\
{\tt \{chuxiangxiang,zhangbo11,xuruijun\}@xiaomi.com}
}
\iccvfinalcopy

\def\iccvPaperID{5476} 

\begin{document}
\maketitle

\begin{abstract}
One of the most critical problems in weight-sharing neural architecture search is the evaluation of candidate models within a predefined search space. In practice, a one-shot supernet is trained to serve as an evaluator. A faithful ranking certainly leads to more accurate searching results. However, current methods are prone to making misjudgments. In this paper, we prove that their biased evaluation is due to inherent unfairness in the supernet training. In view of this, we propose two levels of constraints: \textbf{expectation fairness} and \textbf{strict fairness}. Particularly, strict fairness ensures equal optimization opportunities for all choice blocks throughout the training, which neither overestimates nor underestimates their capacity. We demonstrate that this is crucial for improving the confidence of models' ranking. Incorporating the one-shot supernet trained under the proposed fairness constraints with a multi-objective evolutionary search algorithm, we obtain various state-of-the-art models, e.g., FairNAS-A attains \textbf{77.5\%} top-1 validation accuracy on ImageNet. 
\end{abstract}

\section{Introduction}
The advent of neural architecture search (NAS) has brought deep learning into an era of automation \cite{zoph2016neural}. Abundant efforts have been dedicated to searching  within carefully designed search space \cite{zoph2017learning,real2018regularized,tan2018mnasnet,lu2018nsga,tan2019efficientnet}. 
Meanwhile, the evaluation of a network's performance is an important building block for NAS. Conventional approaches evaluate an enormous amount of models based on resource-devouring training \cite{zoph2017learning,tan2018mnasnet}. Recent attention has been drawn to improve its efficiency via parameter sharing \cite{brock2017smash,liu2018darts,pham2018efficient,wu2018fbnet}. 

Generally speaking, the weight-sharing approaches all involve training a supernet that incorporates many candidate subnetworks. They can be roughly classified into two categories: those who couple searching and training within one stage \cite{pham2018efficient,liu2018darts,cai2018proxylessnas,stamoulis2019single,wu2018fbnet} and others who decouple them into two stages, where the trained supernet is treated as an evaluator for final searching \cite{brock2017smash,bender2018understanding,guo2019single,luo2018neural}. The supernet is a so-called one-shot model.

Despite being widely utilized due to searching efficiency, weight sharing approaches are roughly built on empirical experiments instead of solid theoretical ground. Several fundamental issues remain to be addressed.  Namely, 
\begin{enumerate*}
\item Why is there a large gap between the range of supernet predicted accuracies and that of ``ground-truth" ones by stand-alone training from scratch \cite{brock2017smash,bender2018understanding}? 
\item How to build a good evaluator that neither overestimates nor underestimates subnetworks? 
\item Why does the weight-sharing mechanism work, if under some conditions?
\end{enumerate*}

\begin{figure*}[ht]
\centering
\subfigure{
\includegraphics[width=0.48\textwidth,scale=0.6]{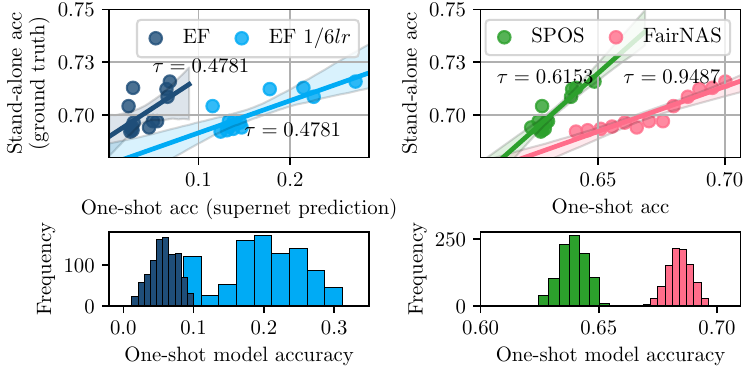}
\includegraphics[width=0.48\textwidth,scale=0.6]{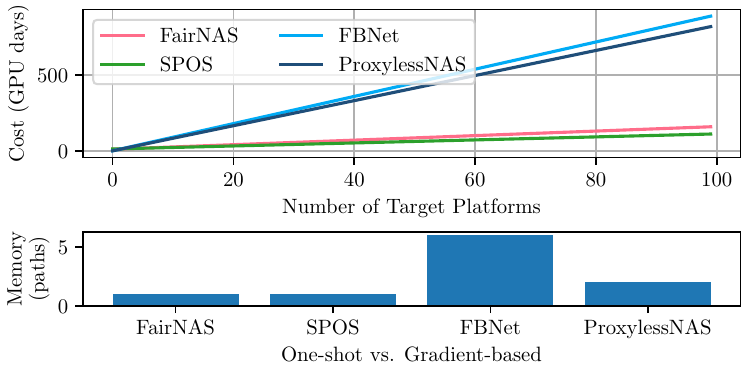}
}
\caption{\textbf{Left:} The supernet trained with Strict Fairness (FairNAS) gives more reliable accuracy prediction (higher correlation $\tau$) than those of Expectation Fairness (EF). \emph{Top:} Relation between supernet-predicted accuracies on ImageNet and ground-truth ones.
\emph{Bottom:} Histograms of validation accuracies from a stratified sample (960 each) of one-shot models.  Note EF baselines sample one path and perform $k=6$ iterations at each step, while SPOS \cite{guo2019single} is a special case of EF. All methods use the same $lr$ except the light blue one that reduces to $1/6 lr$. \textbf{Right:} Comparison of amortized GPU cost and memory consumption} 
\label{fig:fairness-hist}
\end{figure*} 

In this paper, we attempt to answer the above three questions for two-stage weight-sharing approaches. We  present Fair Neural Architecture Search (FairNAS) in which we train the supernet under the proposed fairness constraints to improve evaluation confidence. Our analysis and experiments are conducted in a widely used search space as in \cite{cai2018proxylessnas,wu2018fbnet,guo2019single,stamoulis2019single}, as well as a cell-based search space from a common benchmark NAS-Bench-201 \cite{dong2019bench}. The contributions can be summarized as follows.

\textbf{Firstly}, we prove it is due to \emph{unfair bias} that the supernet misjudges submodels' performance, which is inevitable in current one-shot approaches \cite{brock2017smash,bender2018understanding}.

\textbf{Secondly}, we propose two levels of fairness constraints: \textit{Expectation Fairness} (EF) and  \textit{Strict Fairness} (SF). They are enforced to alleviate supernet bias and to boost evaluation capability. Both outperform the existing unfair approaches while SF performs best with a ranking ($\tau$) of 0.7412 on NAS-Bench-201.

\begin{figure*}[ht]
\centering
\begin{tikzpicture}[every node/.style={scale=0.6},node distance=0.8cm]
\node (a) at (0,3cm)  {\includegraphics[width=2cm,height=2cm]{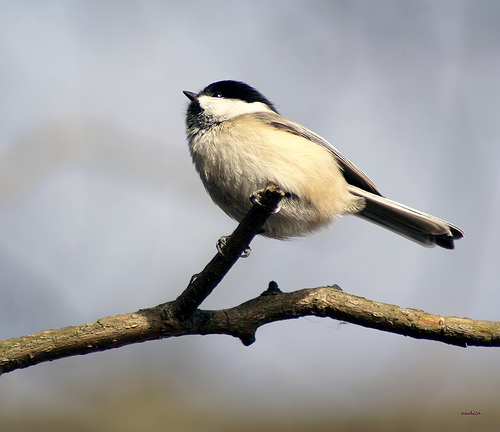}};
\node (op1) [op, fill=xayellow, right of=a, xshift=1.5cm, yshift=2cm] {};
\node (op2) [op, fill=xablue, below of=op1] {};
\node (op3) [op, fill=xapink,below of=op2] {};
\node (op4) [op, fill=xagreen,below of=op3] {};
\node (op5) [op, fill=xagray,below of=op4] {};
\node (op6) [op, fill=salmon,below of=op5] {};
\node [scale=1.4,fit=(op1)(op2)(op3)(op4)(op5)(op6), label={[yshift=-5.2cm]\Large Layer 1}] (l1) {};
\draw [xagray] (l1.north west)+(0,0.3cm) rectangle ([yshift=-0.3cm]l1.south east) ;
\begin{scope}[zlevel=background]
\node[right of=l1,xshift=10.2cm,yshift=-0.15cm] (b){\includegraphics[scale=0.85]{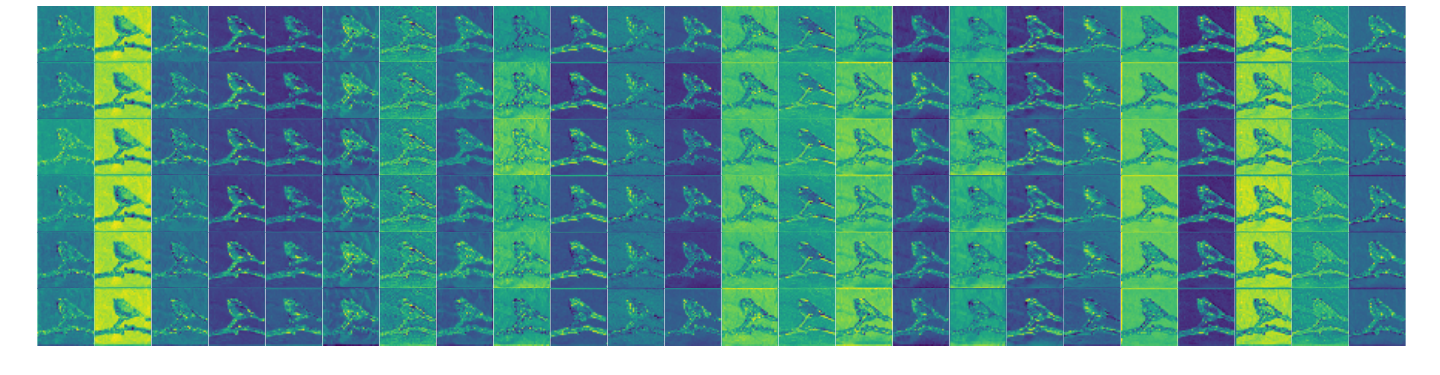}};
\end{scope}
\node[right of=b,xshift=11.8cm,yshift=0.2cm] (b){\includegraphics{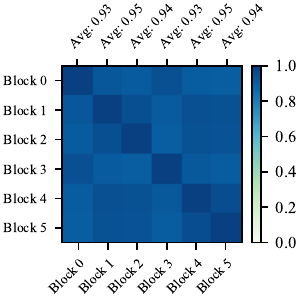}};
\draw [-latex,xagray] (a) to [out=0,in=180]   (op1);
\draw [-latex,xagray] (a) to [out=0,in=180]   (op2);
\draw [-latex,xagray] (a) to [out=0,in=180]   (op3);
\draw [-latex,xagray] (a) to [out=0,in=180]   (op4);
\draw [-latex,xagray] (a) to [out=0,in=180]   (op5);
\draw [-latex,xagray] (a) to [out=0,in=180]   (op6);
\draw [-latex,xagray] (op1) -- +(0.5,0);
\draw [-latex,xagray] (op2) -- +(0.5,0);
\draw [-latex,xagray] (op3) -- +(0.5,0);
\draw [-latex,xagray] (op4) -- +(0.5,0);
\draw [-latex,xagray] (op5) -- +(0.5,0);
\draw [-latex,xagray] (op6) -- +(0.5,0);
\end{tikzpicture}
\caption{\textbf{Left:} Feature maps activated from 6 blocks of the first layer in our supernet trained with \emph{strict fairness}. \textbf{Right:} Cross-block cosine similarity averaged on each channel. Each block learns very similar feature maps (similarity all above 0.9)}
\label{fig:fair_activation}
\end{figure*}

\textbf{Thirdly}, we unveil the root cause of the validity of single-path supernet training under our fairness perspective. That is, different choice blocks are interchangeable during the supernet training as they learn similar feature maps, according to their high \emph{cosine similarity} measure, see Figure~\ref{fig:fair_activation}.

\textbf{Last but not the least}, our fair single-path sampling is memory-friendly, and its GPU costs can also be linearly amortized to the number of target models, see Fig.~\ref{fig:fairness-hist}.  We then incorporate our supernet with an EA-based multi-objective searching framework, from which we obtain three state-of-the-art networks within a single proxyless run at a cost of 12 GPU days on ImageNet. 


\section{Fairness Taxonomy of Weight-sharing NAS} 

\subsection{Review of Biased Supernets}\label{sec:review-bias}
On the one hand, supernet training and searching for good models are \emph{nested}. In ENAS \cite{pham2018efficient}, the sampling policy $\pi(m, \theta)$ of an LSTM controller \cite{hochreiter1997long} and a sampled subnetwork $m$ are alternatively trained. The final models are sampled again by the trained policy $\pi$, one who has the highest reward on a mini-batch of validation data is finally chosen. DARTS \cite{liu2018darts} combines the supernet training and searching within a bi-level optimization where each operation is associated with a coefficient denoting its importance. Both two methods treat all subnetworks unequally and introduce gradually increasing biases through optimization. Those who have better initial performance are more likely to be sampled or to maintain higher coefficients, resulting in a suboptimal or an even worse solution. For instance, architectures from DARTS usually contain an excessive number of skip connections \cite{zela2019understanding,liang2019darts+}, which damage the outcome performance. Therefore, the prior-learning DARTS is biased as per skip connections, while a random approach doesn't suffer \cite{li2019random}. DARTS overrated `bad' models (jammed by skip connections), meantime many other good candidates are depreciated.

On the other hand, the rest one-shot methods consider the trained supernet as a confident proxy, which we also follow, to predict the real performance of all subnetworks \cite{brock2017smash,bender2018understanding,guo2019single}. We emphasize that a reliable proxy supernet should neither severely overestimate nor underestimate the ground-truth performance of any model. The next searching stage is decoupled from training and it can be implemented with random sampling, evolutionary algorithms, or reinforcement learning. 

SMASH \cite{brock2017smash} invents a hyper network (referred to as HyperNet $H$) to generate the weights of a neural architecture by its binary encoding. This HyperNet resembles a typical supernet in that they can both produce weights for any architecture in the search space. At each step, a model is randomly sampled and trained based on the generated weights from $H$, and in turn, it updates the weights of $H$. For a set of randomly sampled models, a correlation between predicted validation errors and ground-truth exists, but it has a large discrepancy between the ranges, i.e., 40\%-90\% vs. 25\%-30\% on CIFAR-100 \cite{krizhevsky2009learning}.

One-Shot \cite{bender2018understanding} involves a dynamic dropout rate for the supernet, each time only a subset is optimized. Apart from its training difficulty, there is also an evident performance gap of submodels with inherited weights compared with their ground-truth, i.e., 30\%-90\% vs. 92\%-94.5\% on CIFAR-10 \cite{krizhevsky2009learning}. 
%

\subsection{Evaluating Supernets by Ranking Ability}
Regardless of how supernets are trained, what matters most is how well they predict the performance of candidate models. To this end, a recent work \cite{sciuto2019evaluating} evaluates weight-sharing supernets with Kendall Tau ($\tau$) metric \cite{kendall1938new}. It measures the relation between one-shot models and stand-alone trained ones. The range of $\tau$ is from -1 to 1, meaning the rankings are totally reversed or completely preserved, whereas 0 indicates no correlation at all. Surprisingly, most recent approaches behave incredibly poorly on this metric \cite{luo2018neural,pham2018efficient,sciuto2019evaluating}. A method based on time-consuming incomplete training only reaches an average $\tau$ of 0.474 \cite{zheng2019multi}.

Given the above biased supernets, we are motivated to revisit one-shot approaches to discover what might be the cause of the obvious range disparity, and, how much does unfair training affect their ranking ability. We ponder that if the supernet is trained free of bias, will it improve evaluation confidence and narrow the accuracy gap? Next, we start with a formal discussion of fairness.

\subsection{Formal Formulation of Fairness}
What kind of fairness can we think of? Will fairness help to improve supernet performance and ranking ability? First of all, to remove the training difference between a supernet and its submodels, we scheme an \emph{equality principle} on training modality.
\begin{definition}\label{def:equal}
\textbf{Equality Principle.} Training a supernet satisfies the \emph{equality principle} if and only if it is in the same way how a submodel is trained.
\end{definition}

Only those who train a single path model at each step meet this principle by its definition. On the contrary, other methods like DARTS \cite{liu2018darts} train the supernet with all paths altogether, One-Shot \cite{bender2018understanding} dynamically drops out some paths,  and ProxylessNAS \cite{cai2018proxylessnas} uses two paths, directly violating the principle.

Formally, we discuss fairness in a common supernet that consists of $L$ layers, each with several choice blocks. Without loss of generality, we suppose each layer has an equal number of choices, say $m$. A model is generated by sampling a block layer by layer. The weights are updated for $n$ times in total. Therefore, we can describe the training process as $P(m, n, L)$. 



\subsubsection{First Attempt: Expectation Fairness}
In order to reduce the above mentioned bias in Section \ref{sec:review-bias}, a natural way is to guarantee all choices blocks have equal expectations after $n$ steps. We define this basic requirement as \emph{expectation fairness} in Definition~\ref{def:ef}.
\begin{definition}
	\label{def:ef}
	\textbf{Expectation Fairness.} On the basis of Definition~\ref{def:equal}, let $\Omega$ be the sampling space containing $m$ basic events $\{l_1, l_2, ..., l_m\}$, which are generated by selecting a block from layer $l$ with $m$ choice blocks.
	Let $Y_{l_i}$ be the number of times that the outcome $l_i$ is observed (updated) over $n$ trials. 
	Then the expectation fairness is that for $P(m, n, L)$, $ E(Y_{l_1})=E(Y_{l_2})=...=E(Y_{l_m})$ holds, $\forall l \in L$.
\end{definition}

\subsubsection{An EF Example: Uniform Sampling}

Let us check a single-path routine \cite{guo2019single} which uses \emph{uniform sampling}. As sampling on any layer $l$ is independent of others, we first consider the case $P(m, n, l)$. Selecting a block from layer $l$ is subject to a categorical distribution. 
In this case, each basic event occurs with an equal probability $p({X=l_i}) = \frac{1}{m}$. For $n$ steps, the expectation and variance of $Y_{l_i}$ can be written as,
\begin{equation}\label{eq:unfair}
\begin{split}
E(Y_{l_i}) & = n * p_{l_i}=n/m \\
\textit{Var}(Y_{l_i}) &=n * p_{l_i}(1-p_{l_i})=\frac{n(m-1)}{m^2}
\end{split}
\end{equation}

That's to say, all choices share the same expectation and variance. Consequently, \textbf{uniform sampling meets Expectation Fairness} by Definition~\ref{def:ef} and it seems superficially fair for various choices. However,  Expectation Fairness is not enough. For example, we can randomly sample each model and keep it training for $k$ times, then switch to another. This procedure also meets Definition~\ref{def:ef}, but it's very unstable to train.

Even in SPOS \cite{guo2019single} with uniform sampling, there is a latent \textbf{ordering issue}. For a sequence of choices ($M_1$, $M_2$, $M_3$), it implies an inherent training order $M_1 \to M_2 \to M_3$. Since each model is usually trained by back-propagation,
the trained weights of $M_1$ are immediately updated to the supernet and those of $M_2$ are renewed next while carrying the effect of the former update, so for $M_3$. A simple permutation of ($M_1$, $M_2$, $M_3$) does comply with Expectation Fairness but yields different results. Besides, if the learning rate $lr$ is changed within the sequence, the situation thus becomes even more complicated. 

Generally, for $P(m, n, L)$ where $m,n,L$ are positive integers,  
assume the sampling times $n$ can be divided by $m (\geq2)$. If we adopt uniform sampling, as $n$ goes infinite, \textbf{it is impossible for $m$ choices to be sampled for an exactly equal number of times}. This is formally stated as below.

\begin{lemma}
	\label{lem:m}
	Regarding $P(m, n, L)$, $\forall$ n  $\in \{x: x \% m = 0,  x \in N_+  \}$, $\lim\limits_{n \to +\infty} p(Y_{l1}=Y_{l2}=...=Y_{lm})=0$.
\end{lemma}
\begin{proof}
	Let $f(m,n) = p(Y_{l1} = Y_{l2}=...=Y_{lm})$.
	\begin{equation}
	f(m,n) =
	C_n^\frac{n}{m}C_\frac{n(m-1)}{m}^\frac{n}{m}...C_\frac{n}{m}^\frac{n}{m}\frac{1}{m^n}=
	\frac{n!}{(\frac{n}{m}!)^m}\frac{1}{m^n}
	\end{equation}
	
	Firstly, we prove the existence of limitation, $f(n)$ strictly decreases monotonically with $n$ and $f(n) \ge 0$, therefore, its limitation exists. 
	
	Secondly, we calculate its limitation using equivalent infinity replacement based on Stirling's approximation about factorial \cite{tweddle2012james}.
	\begin{equation}
	\begin{split}
	\lim\limits_{n \to +\infty} f(m,n) &= \lim\limits_{n \to +\infty}\frac{n!}{(\frac{n}{m}!)^m\times m^n} \\
	&= \lim\limits_{n \to +\infty} \frac{\sqrt{2\pi n}(\frac{n}{e})^n}{\sqrt{2\pi \frac{n}{m}}^m (\frac{n}{e})^n} \\
	&= \lim\limits_{n \to +\infty} \frac{\sqrt{m}}{\frac{2\pi n}{m}^\frac{m-1}{2}} =0\\
	\end{split}
	\end{equation}
	Q.E.D.
\end{proof}

%
%

\begin{figure}[ht]
	\centering
	\includegraphics[scale=0.6]{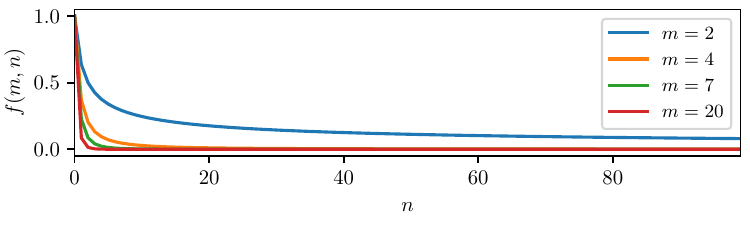}\\
	\caption{The function curve of $f(m,n)$ in Lemma~\ref{lem:m}. When sampling uniformly from $m$ blocks for $n$ trials, the probability of having an equal sampling number for each block quickly reaches zero. }
	\label{fig:f1n}
\end{figure}

Lemma~\ref{lem:m} is somewhat counter-intuitive and thereby neglected in previous works. To throw light on this phenomenon, we plot this probability curve in Figure~\ref{fig:f1n}. We see that $f(2, n)$ decreases below 0.2 when $n \ge 20$.  In most cases, $n \ge 10^6$, which suffers severely from this issue.

\subsubsection{A Meticulous Overhaul: Strict Fairness}

Our insights come from the above overlooked phenomenon. We propose a more rigorous requirement that ensures the parameter of every choice block be updated the same amount of times at any stage, which is called strict fairness and formally as Definition~\ref{def:sf}.

\begin{definition}
	\label{def:sf}
	\textbf{Strict Fairness.} Regarding P(m, n, L), $\forall$ n  $\in \{x: x \% m = 0,  x \in N_+  \}$ , $Y_{l_1}=Y_{l_2}=...=Y_{l_m}$ holds.
\end{definition}
Definition~\ref{def:sf} imposes a constraint more demanding than Definition~\ref{def:ef}. That is, $p(Y_{l_1}=Y_{l_2}=...=Y_{l_m}) = 1$ holds at any time. It seems subtle but it will be later proved to be crucial. Nevertheless, we have to be aware that it is not ultimate fairness since different models have their own optimal initialization strategy and hyperparameters, which we single them out for simplicity.  

\section{Fair Neural Architecture Search}

Our NAS pipeline is divided into two stages: training the supernet and searching for competitive models.

\subsection{Stage 1: Train Supernet with Strict Fairness}
We first propose a fair sampling and training algorithm to strictly abide by Defintion~\ref{def:sf}.  We use \emph{uniform sampling without replacement} and sample $m$ models at step $t$ so that each choice block must be activated and updated only once. This is detailed in Algorithm~\ref{alg:fair_sample} and depicted by Figure \ref{fig:one-shot-strategy}. 

\begin{algorithm}
	\caption{\textbf{: Stage 1} - Fair Supernet Training.}
	\label{alg:fair_sample}
	\begin{algorithmic}
		\STATE {\bfseries Input:} training steps $n$, search space $S_{(m,L)}$, $m\times L$ supernet parameters $\Theta{(m,L)}$, search layer depth $L$, choice blocks $m$ per layer, training epochs $N$, training data loader $D$, loss function $Loss$
		\STATE initialize  every $\theta_{j,l}$ in  $\Theta_{(m,L)}$.
		\FOR{$i=1$ {\bfseries to} $N$}
		\FOR{$data,labels$ {\bfseries in} $D$}
		\FOR{$l=1$ {\bfseries to} $L$}
		\STATE {$c_l$ = an uniform index permutation for the choices of layer $l$} 
		\ENDFOR
		\STATE  Clear gradients recorder for all parameters
		\STATE $\nabla \theta_{j, l} =0, j=1,2,...,m, l=1,2,...,L$ 
		\FOR{$k=1$ {\bfseries to} $m$}
		\STATE Build $model_k = (c_{1_k},c_{2_k},..,c_{L_k})$ from sampled index
		\STATE Calculate gradients for $model_k$ based on $Loss$, data, labels.
		\STATE Accumulate gradients for activated parameters, $\nabla \theta_{c_{1_k},1}, \nabla \theta_{c_{2_k},2},..., \nabla \theta_{c_{L_k},L}$
		\ENDFOR
		\STATE update $\theta_{(m,L)}$ by accumulated gradients. 
		\ENDFOR
		\ENDFOR
	\end{algorithmic}
\end{algorithm}

\begin{figure*}[ht]
	\centering
	\begin{tikzpicture}[thick,scale=0.6, every node/.style={scale=0.6,minimum width=3cm,font=\Large},node distance=1cm]
	\node (s0) [darklayer] at (0,0) {\textcolor{white}{stem}};
	\node (1) [cell, below of=s0] {Layer 1};
	\node (2) [cell, below of=1] {Layer 2};
	\node (3) [below of=2] {...};
	\node (4) [cell, below of=3] {Layer L};
	\node (t0) [darklayer, below of=4] {\textcolor{white}{tail}};
	\draw [arrow] (s0) -- (1);
	\draw [arrow] (1) -- (2);
	\draw [arrow] (2) -- (3);
	\draw [arrow] (3) -- (4);
	\draw [arrow] (4) -- (t0);
	\draw [arrow] (t0.south) -- +(0,-0.3cm);
	\draw [arrow] (s0.north)+(0,0.3cm) -- (s0.north);
	\node (a) [right of=s0, xshift=1cm,yshift=1cm] {};
	\node (b) [right of=t0, xshift=1cm,yshift=-0.5cm] {};
	\draw [dotted] (a) -- (b);
	\begin{scope}[xshift=4cm,yshift=-1cm,thin] 
	\node (l1op1) [op, fill=xayellow,label={[label distance=.1cm]op1}]  {};
	\node (l1op2) [op, fill=xablue!20,label={[label distance=.1cm]op2},right of=l1op1] {};
	\node (l1op3) [op, fill=xapink!20,label={[label distance=.1cm]op3},right of=l1op2] {};
	\node (l1op4) [op, fill=xagreen!20,label={[label distance=.1cm]op4},right of=l1op3] {};
	\node [scale=1.3,fit=(l1op1)(l1op2)(l1op3)(l1op4)] (l1) {};
	\begin{pgfonlayer}{background}	
	\filldraw [fit=(l1),fill=gray!0,draw=xagray](l1.south -| l1.west) rectangle (l1.north -| l1.east);
	\end{pgfonlayer}
	\node (l2op1) [op, fill=xayellow!20, below of=l1op1] {};
	\node (l2op2) [op, fill=xablue!20, right of=l2op1] {};
	\node (l2op3) [op, fill=xapink,right of=l2op2] {};
	\node (l2op4) [op, fill=xagreen!20,right of=l2op3] {};
	\node [scale=1.3,fit=(l2op1)(l2op2)(l2op3)(l2op4)] (l2) {};
	\begin{pgfonlayer}{background}	
	\filldraw [fit=(l2),fill=gray!0,draw=xagray](l2.south -| l2.west) rectangle (l2.north -| l2.east);
	\end{pgfonlayer}
	\node (liop1) [below of=l2op1]{};
	\node (liop2) [right of=liop1,xshift=0.5cm]{...};
	\node (l3op1) [op, fill=xayellow!20, below of=liop1] {};
	\node (l3op2) [op, fill=xablue!20, right of=l3op1] {};
	\node (l3op3) [op, fill=xapink!20, right of=l3op2] {};
	\node (l3op4) [op, fill=xagreen,right of=l3op3] {};
	\node [scale=1.3,fit=(l3op1)(l3op2)(l3op3)(l3op4)] (l3) {};
	\begin{pgfonlayer}{background}	
	\filldraw [fit=(l3),fill=gray!0,draw=xagray](l3.south -| l3.west) rectangle (l3.north -| l3.east);
	\end{pgfonlayer}
	\draw [arrow] (l1op1) -- (l2op3);
	\draw [arrow] (l2op3) -- (liop2);
	\draw [arrow] (liop2) -- (l3op4);
	\end{scope}
	\begin{scope}[xshift=9cm,yshift=-1cm,thin]
	\node (l1op1) [op, fill=xayellow!20,label={[label distance=.1cm]op1}] {};
	\node (l1op2) [op, fill=xablue!20,label={[label distance=.1cm]op2},right of=l1op1] {};
	\node (l1op3) [op, fill=xapink,label={[label distance=.1cm]op3},right of=l1op2] {};
	\node (l1op4) [op, fill=xagreen!20,label={[label distance=.1cm]op4},right of=l1op3] {};
	\node [scale=1.3,fit=(l1op1)(l1op2)(l1op3)(l1op4)] (l1) {};
	\begin{pgfonlayer}{background}	
	\filldraw [fit=(l1),fill=gray!0,draw=xagray](l1.south -| l1.west) rectangle (l1.north -| l1.east);
	\end{pgfonlayer}
	\node (l2op1) [op, fill=xayellow, below of=l1op1] {};
	\node (l2op2) [op, fill=xablue!20, right of=l2op1] {};
	\node (l2op3) [op, fill=xapink!20,right of=l2op2] {};
	\node (l2op4) [op, fill=xagreen!20,right of=l2op3] {};
	\node [scale=1.3,fit=(l2op1)(l2op2)(l2op3)(l2op4)] (l2) {};
	\begin{pgfonlayer}{background}	
	\filldraw [fit=(l2),fill=gray!0,draw=xagray](l2.south -| l2.west) rectangle (l2.north -| l2.east);
	\end{pgfonlayer}
	\node (liop1) [below of=l2op1]{};
	\node (liop2) [right of=liop1,xshift=0.5cm]{...};
	\node (l3op1) [op, fill=xayellow!20, below of=liop1] {};
	\node (l3op2) [op, fill=xablue!20, right of=l3op1] {};
	\node (l3op3) [op, fill=xapink, right of=l3op2] {};
	\node (l3op4) [op, fill=xagreen!20,right of=l3op3] {};
	\node [scale=1.3,fit=(l3op1)(l3op2)(l3op3)(l3op4)] (l3) {};
	\begin{pgfonlayer}{background}	
	\filldraw [fit=(l3),fill=gray!0,draw=xagray](l3.south -| l3.west) rectangle (l3.north -| l3.east);
	\end{pgfonlayer}
	\draw [arrow] (l1op3) -- (l2op1);
	\draw [arrow] (l2op1) -- (liop2);
	\draw [arrow] (liop2) -- (l3op3);
	\end{scope} 
	\begin{scope}[xshift=14cm,yshift=-1cm,thin]
	\node (l1op1) [op, fill=xayellow!20,label={[label distance=.1cm]op1}] {};
	\node (l1op2) [op, fill=xablue,label={[label distance=.1cm]op2},right of=l1op1] {};
	\node (l1op3) [op, fill=xapink!20,label={[label distance=.1cm]op3},right of=l1op2] {};
	\node (l1op4) [op, fill=xagreen!20,label={[label distance=.1cm]op4},right of=l1op3] {};
	\node [scale=1.3,fit=(l1op1)(l1op2)(l1op3)(l1op4)] (l1) {};
	\begin{pgfonlayer}{background}	
	\filldraw [fit=(l1),fill=gray!0,draw=xagray](l1.south -| l1.west) rectangle (l1.north -| l1.east);
	\end{pgfonlayer}
	\node (l2op1) [op, fill=xayellow!20, below of=l1op1] {};
	\node (l2op2) [op, fill=xablue!20, right of=l2op1] {};
	\node (l2op3) [op, fill=xapink!20,right of=l2op2] {};
	\node (l2op4) [op, fill=xagreen,right of=l2op3] {};
	\node [scale=1.3,fit=(l2op1)(l2op2)(l2op3)(l2op4)] (l2) {};
	\begin{pgfonlayer}{background}	
	\filldraw [fit=(l2),fill=gray!0,draw=xagray](l2.south -| l2.west) rectangle (l2.north -| l2.east);
	\end{pgfonlayer}
	\node (liop1) [below of=l2op1]{};
	\node (liop2) [right of=liop1,xshift=0.5cm]{...};
	\node (l3op1) [op, fill=xayellow, below of=liop1] {};
	\node (l3op2) [op, fill=xablue!20, right of=l3op1] {};
	\node (l3op3) [op, fill=xapink!20, right of=l3op2] {};
	\node (l3op4) [op, fill=xagreen!20,right of=l3op3] {};
	\node [scale=1.3,fit=(l3op1)(l3op2)(l3op3)(l3op4)] (l3) {};
	\begin{pgfonlayer}{background}	
	\filldraw [fit=(l3),fill=gray!0,draw=xagray](l3.south -| l3.west) rectangle (l3.north -| l3.east);
	\end{pgfonlayer}
	\draw [arrow] (l1op2) -- (l2op4);
	\draw [arrow] (l2op4) -- (liop2);
	\draw [arrow] (liop2) -- (l3op1);
	\end{scope} 
	\begin{scope}[xshift=19cm,yshift=-1cm,thin]
	\node (l1op1) [op, fill=xayellow!20,label={[label distance=.1cm]op1}] {};
	\node (l1op2) [op, fill=xablue!20,label={[label distance=.1cm]op2},right of=l1op1] {};
	\node (l1op3) [op, fill=xapink!20,label={[label distance=.1cm]op3},right of=l1op2] {};
	\node (l1op4) [op, fill=xagreen,label={[label distance=.1cm]op4},right of=l1op3] {};
	\node [scale=1.3,fit=(l1op1)(l1op2)(l1op3)(l1op4)] (l1) {};
	\begin{pgfonlayer}{background}	
	\filldraw [fit=(l1),fill=gray!0,draw=xagray](l1.south -| l1.west) rectangle (l1.north -| l1.east);
	\end{pgfonlayer}
	\node (l2op1) [op, fill=xayellow!20, below of=l1op1] {};
	\node (l2op2) [op, fill=xablue, right of=l2op1] {};
	\node (l2op3) [op, fill=xapink!20,right of=l2op2] {};
	\node (l2op4) [op, fill=xagreen!20,right of=l2op3] {};
	\node [scale=1.3,fit=(l2op1)(l2op2)(l2op3)(l2op4)] (l2) {};
	\begin{pgfonlayer}{background}	
	\filldraw [fit=(l2),fill=gray!0,draw=xagray](l2.south -| l2.west) rectangle (l2.north -| l2.east);
	\end{pgfonlayer}
	\node (liop1) [below of=l2op1]{};
	\node (liop2) [right of=liop1,xshift=0.5cm]{...};
	\node (l3op1) [op, fill=xayellow!20, below of=liop1] {};
	\node (l3op2) [op, fill=xablue, right of=l3op1] {};
	\node (l3op3) [op, fill=xapink!20, right of=l3op2] {};
	\node (l3op4) [op, fill=xagreen!20,right of=l3op3] {};
	\node [scale=1.3,fit=(l3op1)(l3op2)(l3op3)(l3op4)] (l3) {};
	\begin{pgfonlayer}{background}	
	\filldraw [fit=(l3),fill=gray!0,draw=xagray](l3.south -| l3.west) rectangle (l3.north -| l3.east);
	\end{pgfonlayer}
	\draw [arrow] (l1op4) -- (l2op2);
	\draw [arrow] (l2op2) -- (liop2);
	\draw [arrow] (liop2) -- (l3op2);
	\end{scope}
	\begin{scope}[xshift=24cm,yshift=-1cm,thin]
	\node (l1op1) [op, fill=xayellowd,label={[label distance=.1cm]op1}] {};
	\node (l1op2) [op, fill=xablued,label={[label distance=.1cm]op2},right of=l1op1] {};
	\node (l1op3) [op, fill=xapinkd,label={[label distance=.1cm]op3},right of=l1op2] {};
	\node (l1op4) [op, fill=xagreend,label={[label distance=.1cm]op4},right of=l1op3] {};
	\node [scale=1.3,fit=(l1op1)(l1op2)(l1op3)(l1op4)] (l1) {};
	\begin{pgfonlayer}{background}	
	\filldraw [fit=(l1),fill=gray!0,draw=xagray](l1.south -| l1.west) rectangle (l1.north -| l1.east);
	\end{pgfonlayer}
	\node (l2op1) [op, fill=xayellowd, below of=l1op1] {};
	\node (l2op2) [op, fill=xablued, right of=l2op1] {};
	\node (l2op3) [op, fill=xapinkd,right of=l2op2] {};
	\node (l2op4) [op, fill=xagreend,right of=l2op3] {};
	\node [scale=1.3,fit=(l2op1)(l2op2)(l2op3)(l2op4)] (l2) {};
	\begin{pgfonlayer}{background}	
	\filldraw [fit=(l2),fill=gray!0,draw=xagray](l2.south -| l2.west) rectangle (l2.north -| l2.east);
	\end{pgfonlayer}
	\node (liop1) [below of=l2op1]{};
	\node (liop2) [right of=liop1,xshift=0.5cm]{...};
	\node (l3op1) [op, fill=xayellowd, below of=liop1] {};
	\node (l3op2) [op, fill=xablued, right of=l3op1] {};
	\node (l3op3) [op, fill=xapinkd, right of=l3op2] {};
	\node (l3op4) [op, fill=xagreend,right of=l3op3] {};
	\node [scale=1.3,fit=(l3op1)(l3op2)(l3op3)(l3op4)] (l3) {};
	\begin{pgfonlayer}{background}	
	\filldraw [fit=(l3),fill=gray!0,draw=xagray](l3.south -| l3.west) rectangle (l3.north -| l3.east);
	\end{pgfonlayer}
	\end{scope}
	\node (st) [right of=s0, xshift=2cm,yshift=0.5cm] {step $t$:};  
	\draw [decorate,decoration={brace,amplitude=5pt}] (st.south east) --+(17cm,0) node [midway,yshift=0.5cm,above] {sample $m$ models without replacement and train them sequentially}; 
	\node (sn1) at (6.2,-3) {};
	\node [right of=sn1,xshift=4cm] (sn2) {};
	\node [right of=sn2,xshift=4cm] (sn3) {};
	\node [right of=sn3,xshift=4cm] (sn4) {};
	\draw [arrow] (sn1) --node[above]{$\nabla \theta_1$}  +(2cm,0);
	\draw [arrow] (sn2) --node[above]{$\nabla \theta_2$} +(2cm,0);
	\draw [arrow] (sn3) --node[above]{$\nabla \theta_3$} +(2cm,0);
	\draw [arrow] (sn4) --node[above]{$\nabla \theta_4$}  node[below]{ update $\theta$}+(2cm,0);
	\end{tikzpicture}
	\caption[]{Our \emph{strict fairness sampling and training} strategy for supernet. A supernet training step $t$ consists of training $m$ models, each on one batch of data.  The supernet gets its weights updated after accumulating gradients from each single-path model. All operations are thus ensured to be equally sampled and trained within every step $t$. There are  $(6!)^{18}$ choices per step in our experiments\footnotemark} 
	\label{fig:one-shot-strategy}
\end{figure*}

To reduce the bias from different training orders, we don't perform back-propagation and update parameters immediately for each model as in the previous works \cite{bender2018understanding,guo2019single}. Instead, we define one \emph{supernet step} as several back-propagation operations (BPs) accompanied by a single parameter update. In particular, given a mini-batch of training data, each of $m$ single-path models is trained with back-propagation. Gradients are then accumulated across the selected $m$ models but supernet's parameters get updated only when all $m$ BPs are done.  This approach also doesn't suffer from the \emph{ordering issue} as each choice block is updated regardless of external learning rate strategies.

\textbf{Strict Fairness Analysis.}
We now check whether our proposed Algorithm~\ref{alg:fair_sample} satisfies Strict Fairness. By its design, each choice block is activated only once during a parameter update step. Thus $Y'_{l_1}=Y'_{l_2}=...=Y'_{l_m}$ holds. In particular, $Y'_{l_1}=Y'_{l_2}=...=Y'_{l_m}=n/m$ holds\footnote{We use $n$ to denote the total number of BPs operations to match Eq~\ref{eq:unfair}.}. Here, we write its expectation and variance as follows:
\begin{equation}
\begin{split}
E(Y'_{l_i}) = n/m, \textit{Var}(Y'_{l_i}) =0
\end{split}
\end{equation}
Compared with Equation~\ref{eq:unfair}, the obvious difference lies in the variance. For the single-path approach with uniform sampling \cite{guo2019single}, the variance spreads along with $n$, which gradually increases the bias. However, our approach calibrates this inclination and assures fairness at every step.

\subsection{Stage 2: Searching with Supernet}\label{sec:search}

For searching, we can either choose random search, vanilla evolutionary algorithms, or reinforcement learning. In practice, there are many requirements and objectives to achieve, e.g., inference time, multiply-adds, and memory costs, etc. This leads us to adopt a multi-objective solution. Besides, as the search space is too vast to enumerate all models, we need an efficient approach to balance the exploration and exploitation trade-off instead of a random sampling strategy. Here we adopt a searching algorithm from an NSGA-II \cite{deb2002fast} with a small variation by using Proximal Policy Optimization \cite{schulman2017proximal} as the default reinforcing algorithm.  The whole process is given in Algorithm~\ref{alg:nas_pipeline}.
Benefiting from the fast evaluation of the weight-sharing supernet, we can achieve tremendous speed-up in terms of GPU days by two orders of magnitudes.  

\begin{algorithm}[tb]
	\caption{\textbf{: Stage 2} - Search Strategy.}
	\label{alg:nas_pipeline}
	\begin{algorithmic}
		\STATE {\bfseries Input:} Supernet $SN$, the number of generations $G$, validation dataset $D$
		\STATE {\bfseries Output: } A set of $K$ individuals on the Pareto front.
		\STATE {Train supernet $SN$ with Algorithm~\ref{alg:fair_sample}.}
		\STATE {Uniform initialization for the populations $P_1$ and $Q_1$.}
		\FOR {$i=1$ {\bfseries to} $G$}
		\STATE{$R_i = P_i \cup Q_i$}
		\FORALL { $p \in R_i$}
		\STATE {Evaluate model $p$ with inherited weights from $SN$ on $D$}
		\ENDFOR
		\STATE {$F = \text{non-dominated-sorting}(R_i)$}
		\STATE{Pick $N$ individuals to form $P_{i+1}$ by ranks and the crowding distance.}
		\STATE {$ M = \text{tournament-selection}(P_{i+1})$}
		\STATE {$Q_{i+1} = \text{crossover}(M) \cup \text{hierarchical-mutation}(M)  $ }
		\ENDFOR
		\STATE {Select $K$ evenly-spaced models from  $P_{G+1}$ to  train}
	\end{algorithmic}
\end{algorithm}

\footnotetext{It can be calculated by $(6^{19} * 5^{19} * 4^{19} * 3^{19} * 2^{19} *1)/6! = (6!)^{18}$.}

\section{Experiments}
\subsection{Setup}\label{ex:v2}
\textbf{Search Space.}
We use several search spaces in this paper. The first one is that of NAS-Bench-201 \cite{dong2019bench}, which is a new benchmark for NAS methods. Besides,
We adopt two extra search spaces to compare with other NAS methods on ImageNet, \textbf{(a)}
A search space for fairness analysis, which is based on MobileNetV2's inverted bottleneck blocks as done in \cite{cai2018proxylessnas}. In particular, we retain the same amount of layers with standard MobileNetV2 \cite{sandler2018mobilenetv2}. Convolution kernels are with the size in (3, 5, 7) and expansion rates are of (3, 6). We keep the number of filters unchanged. Besides, the squeeze-and-excitation block \cite{hu2018squeeze} is excluded.  In total, it has a size of $6^{16}$.   \textbf{(b)} A search space of 19 layers as ProxylessNAS \cite{cai2018proxylessnas}, whose size spreads to $6^{19}$. This is to be on par with various state-of-the-art methods.

\textbf{Training Hyperparameters.}
For NAS-Bench-201, we train the supernet for 50 epochs using a batch size of 128. The initial learning rate is 0.025 and decayed to zero by the cosine schedule.

For search space (a), we train the supernet for 150 epochs using a batch size of 256 and adopt a stochastic gradient descent optimizer with a momentum of 0.9 \cite{sutskever2013importance} based on standard data augmentation as \cite{sandler2018mobilenetv2}. A cosine learning rate decay strategy \cite{loshchilov2016sgdr} is applied with an initial learning rate of 0.045.  Moreover, We regularize the training with L2 weight decay ($4\times10^{-5}$).  Our supernet is thus trained to fullness in 10 GPU days. 

For search space (b), we follow the same strategy as above for training the supernet, but we adopt vanilla data processing as well as training tricks in \cite{tan2018mnasnet} for stand-alone models. Regarding the stand-alone training of sampled models, we use  similar training tricks. To be consistent with the previous works, we don't employ tricks like  cutout \cite{devries2017improved} or mixup \cite{zhang2018mixup}, although they can further improve the scores on the test set.

\subsection{Search Result}
\label{sec:sota}

\subsubsection{Search on ImageNet.}
Figure~\ref{fig:fairnas-a} (supplementary) exhibits the resulting FairNAS-A, B and C models, which are sampled from our Pareto front to meet different hardware constraints. The quantitative result is shown in Table~\ref{tab:comparison-imagenet}.


Notably, FairNAS-A obtains a highly competitive result \textbf{75.3\%} top-1 accuracy for ImageNet classification, which surpasses MnasNet-92 (+0.5\%) and Single-Path-NAS (+0.3\%). FairNAS-B matches Proxyless-GPU with much fewer parameters and multiply-adds. Besides, it surpasses Proxyless-R Mobile (+0.5\%) with a comparable amount of multiply-adds.

Our models also reach a new state of the art when equipped with combined tricks such as Squeeze-and-Excitation \cite{hu2018squeeze}, Swish activations \cite{ramachandran2017searching} and AutoAugment \cite{cubuk2018autoaugment}. Namely, FairNAS-A obtains \textbf{77.5\%} top-1 accuracy by using similar FLOPS as EfficientNet-B0 . Even without mixed kernels \cite{tan2020mixconv}, FairNAS-B ($77.2\%$) outperforms MixNet-M ($77.0\%$) with 11M fewer FLOPS. The most light-weight model FairNAS-C ($76.7\%$)  also outperforms EfficientNet-B0 with about $20\%$ fewer FLOPS.

\begin{table}
	\setlength{\tabcolsep}{2pt}
	\centering
	  \begin{small}
		\begin{tabular}{|l|l|l|l|Hl|r|} 	
		\hline 
			Models & $\times+$  & $P$  &  Top-1  &Top-5 & $M$ & Cost \\
			& (M) & (M) & (\%) & (\%)  & & \scriptsize{(GPU days)}  \\
			\hline
			MobileNetV2 \cite{sandler2018mobilenetv2}   & 300 & 3.4  & 72.0 & 91.0 & - & - \\ 
			\hline 
			NASNet-A \cite{zoph2017learning}  & 564 & 5.3 &74.0 & 91.6 &  SM &1800 \\ 
    		        MnasNet-92 \cite{tan2018mnasnet}  & 388 & 3.9  & 74.8 & 92.1 & SM & $\approx$4k \\ 
			DARTS  \cite{liu2018darts} & 574 & 4.7& 73.3 & 91.3 & SN & 0.5 \\ 
			PC-DARTS (CIFAR10) \cite{Xu2020PC-DARTS:} & 586 & 5.3 & 74.9 & 92.2 & SN & 0.1 \\
			One-Shot Small (F=32) \cite{bender2018understanding} &- & 5.1 & 74.2 & - & SN & 4 \\
		    AtomNAS-A \cite{Mei2020AtomNAS:} & 258 & 3.9 & 74.6 & 92.1 & SN &20.5 \\
			FBNet-B \cite{wu2018fbnet}   & 295 & 4.5 & 74.1 & 91.7 & SP &9 \\  
			Proxyless-R \cite{cai2018proxylessnas} & 320$^\star$ & 4.0  & 74.6 & 92.2 & TP & 8.3 \\ 
			Proxyless GPU \cite{cai2018proxylessnas}  & 465$^\star$ & 7.1 & 75.1 & 92.3 & TP &  8.3 \\ 
		    Single Path One-Shot \cite{guo2019single}  & 323 & 3.5 &  74.4 & 91.8  & SP & 12 \\
			Single-Path NAS \cite{stamoulis2019single} & 365 & 4.3 & 75.0 & 92.2 & SP &1.25 \\ 
			 \textbf{FairNAS-A} (Ours)  & 388& 4.6& \textbf{75.3 } & \textbf{92.4} & SP &12$^\diamond$ \\
			 FairNAS-B (Ours)   & 345 & 4.5 &  75.1 &  92.3 & SP & 12$^\diamond$\\
			 FairNAS-C (Ours)   & 321 & 4.4 &  74.7 & 92.1 & SP & 12$^\diamond$ \\
			 \hline 
			 MnasNet-A2 \cite{tan2018mnasnet}  & 340 & 4.8 & 75.6 & 92.7 & SM & $\approx$4k \\
			 MobileNetV3 Large \cite{howard2019searching} & 219 & 5.4 &75.2 & 92.2 &  SM &$\approx$3k  \\
			 GhostNet 1.3$\times$ \cite{han2020ghostnet} & 226 & 7.3 & 75.7 & 92.7 & - & - \\
			 EfficientNet B0 \cite{tan2019efficientnet} &390 & 5.3  & 76.3 & 93.2 & SM &$\approx$3k\\
			 OFA w/ PS \#75 \cite{cai2019once} & 230 & - & 76.9 & & SP & 175 \\
			 BigNAS-S \cite{yu2020bignas} &  242 & 4.5 & 76.5 & - & SP & ~48$^{\ddagger}$ \\
			 MixNet-M \cite{tan2020mixconv} &360 & 5.0 & 77.0 & 93.3 & SM &$\approx$3k\\
			 AtomNAS-C+ \cite{Mei2020AtomNAS:} & 363 & 5.9 & 77.6 & 93.6 & SN & 20.5 \\
			 \textbf{FairNAS-A}$^{\dagger}$ (Ours) & 392& 5.9& 77.5& \textbf{93.7} &  SP & 12$^\diamond$ \\ 
			 FairNAS-B$^{\dagger}$ (Ours) & 349& 5.7 &77.2 & 93.5 &  SP & 12$^\diamond$  \\
			 FairNAS-C$^{\dagger}$ (Ours) & 325& 5.6 &76.7 & 93.3 & SP & 12$^\diamond$ \\
			\hline
			\noalign{\smallskip\smallskip}
		\end{tabular}
		\end{small}
	\caption{Comparison of mobile models on ImageNet. $M$: Memory cost at all sampled sub-models (SM), a single path or two paths (SP/TP), and a whole supernet (SN). $^\star$: from code,  $^{\dagger}$: w/ SE and Swish, P: Number of parameters, $^\diamond$: Cost shared among A, B and C. $^{\ddagger}$: reportedly 3$\times$ EfficientNet Training } 
	\label{tab:comparison-imagenet}
\end{table}



\subsubsection{Search on NAS-Bench-201.}
To be comparable with existing methods, we formulate our problem as a single objective one: finding the best model. Specially, we use a standard evolutionary algorithm in the second stage after the supernet is fairly trained. The result is shown in Table~\ref{tab:bench201}. Our method outperforms the other baselines in most datasets with the lowest search cost.  

\begin{table*}[ht]
	\small
	\setlength{\tabcolsep}{2pt}
	\centering
	\begin{tabular}{|l|r|*{6}{c|}}
	\hline
Method	& Cost  & \multicolumn{2}{c|}{CIFAR-10}  & \multicolumn{2}{c|}{CIFAR-100}   &  \multicolumn{2}{c|}{ImageNet16-120}  \\
\cline{3-8}
  & (seconds) & valid & test & valid & test & valid & test \\
 \hline
DARTS \cite{liu2018darts} & 11625& 39.77$\pm$0.00 & 54.30$\pm$0.00 & 15.03$\pm$0.00 & 15.61$\pm$0.00 & 16.43$\pm$0.00 & 16.32$\pm$0.00 \\
ENAS \cite{pham2018efficient} & 14058 & 37.51$\pm$3.19 & 53.89$\pm$0.58 & 13.37$\pm$2.35 & 13.96$\pm$2.33 & 15.06$\pm$1.95 & 14.84$\pm$2.10 \\
SETN \cite{dong2019one} & 34139& 84.04$\pm$0.28 & 87.64$\pm$0.00 & 58.86$\pm$0.06 & 59.05$\pm$0.24 & 33.06$\pm$0.02 & 32.52$\pm$0.21   \\ 
GDAS \cite{dong2019search} & 31609 & 89.89$\pm$0.08 & \textbf{93.61$\pm$0.09} & \textbf{71.34$\pm$0.04} & 70.70$\pm$0.30 & 41.59$\pm$1.33 & 41.71$\pm$0.98 \\
FairNAS (ours) & 9845 & \textbf{90.07$\pm$0.57} & 93.23$\pm$0.18 & 70.94$\pm$0.94 & \textbf{71.00$\pm$1.46} & \textbf{41.90$\pm$1.00} & \textbf{42.19$\pm$0.31} \\
\hline
\noalign{\smallskip}
\end{tabular}
	\caption{Comparison on NAS-Bench-201 \cite{dong2019bench}. Averaged on three runs of searching}
\label{tab:bench201}
\end{table*}

\subsection{Transferred Results on CIFAR}
To validate transferability of FairNAS models, we adapt the  pre-trained models on ImageNet  to CIFAR-10 and CIFAR-100 following the configuration of GPipe \cite{huang2018gpipe} and \cite{kornblith2019better}. Table~\ref{table:fairnas-transfer-cifar10} shows that FairNAS models outperform the rest transferred models with higher top-1 accuracy.

\begin{table}
\setlength{\tabcolsep}{2pt}
\small
	\centering
		\begin{tabular}{|l|r|Hr|c|r|c|}
			\hline
			Models & Input Size & Params & \multicolumn{2}{c|}{CIFAR-10}  & \multicolumn{2}{c|}{CIFAR-100} \\
			& &  & $\times+$ & Acc  & $\times+$ & Acc \\
			& & &  \scriptsize{(M)} & \scriptsize{(\%)} & \scriptsize{(M)} & \scriptsize{(\%)} \\
			\hline 
			NASNet-A Large  \cite{zoph2017learning} & 331$\times$331 & 85 &  12030 & 98.0 & 12031 & 86.7$^\star$\\
			EfficientNet-B0 \cite{tan2019efficientnet} & 224$\times$224 & 4.0& 387 &98.1& 387 & 86.8$^\star$\\
			MixNet-M \cite{tan2020mixconv} &224$\times$224& 3.5 & 359 & 97.9 & 359 & 87.1$^\star$ \\
			\textbf{FairNAS-A$^{\dagger}$}  &224$\times$224  & 5.8 & 391 &  \textbf{98.2} & 391 & \textbf{87.3} \\
			FairNAS-B$^{\dagger}$ & 224$\times$224 & 5.6 &  348&98.1 & 348 & 87.0\\
			FairNAS-C$^{\dagger}$ &  224$\times$224 & 5.4 & 324 & 98.0 & 324 & 86.7 \\
			\hline 
			\noalign{\smallskip}
		\end{tabular}
	\caption{Comparison of state-of-the-art methods on CIFAR. $^{\dagger}$: w/ SE and Swish. $^\star$ based on  our reimplementation}
	\label{table:fairnas-transfer-cifar10}
\end{table}
\setlength{\tabcolsep}{1.4pt}

\subsection{Transferability on Object Detection}
For object detection, we treat FairNAS models as drop-in replacements for RetinaNet's backbone \cite{lin2017focal}. We follow the same setting as \cite{lin2017focal} and exploit MMDetection toolbox \cite{chen2019mmdetection} for training. All the models are trained and evaluated on MS COCO dataset (train2017 and val2017 respectively) \cite{lin2014coco} for 12 epochs with a batch size of 16. The initial learning rate is 0.01 and decayed by 0.1$\times$ at epochs 8 and 11.   

The input features from these backbones to the FPN module  are from the last depthwise layers of stage 2 to 5\footnote{We follow the typical nomination for the definition of stages and the orders start from 1.}.  The number of  output channels of FPN is kept 256 as \cite{lin2017focal}. We also use $\alpha=0.25$ and $\gamma=2.0$ for the focal loss. Given longer training epochs and other tricks, the detection performance can be improved further. However, it's sufficient to compare the transferability of various methods. The results are given in Table~\ref{table:fairnas-coco-retina}, we have the best transferability.

\begin{table}[ht]
	\setlength{\tabcolsep}{1pt}
	\small
	\centering
		\begin{tabular}{*{2}{|l}H*{7}{l|}}
			\hline
			Backbones & $\times +$  & Params (M) &Acc    & AP & AP$_{50}$ & AP$_{75}$ & AP$_S$ & AP$_M$ & AP$_L$ \\
			& (M) & (M) & (\%) &(\%) & (\%)& (\%)&(\%) &(\%) &(\%) 
			\\
			\hline 
			MobileNetV2 \cite{sandler2018mobilenetv2} & 300 & 3.4& 72.0 & 28.3 & 46.7 & 29.3 & 14.8 & 30.7 & 38.1\\
			SingPath NAS \cite{stamoulis2019single} & 365 & 4.3 & 75.0 & 30.7 & 49.8 & 32.2 & 15.4 &33.9 & 41.6\\
			MobileNetV3 \cite{howard2019searching} & 219 & & 75.2& 29.9 & 49.3 & 30.8 & 14.9 & 33.3 & 41.1\\
			MnasNet-A2 \cite{tan2018mnasnet} & 340& 4.8 & 75.6 & 30.5 & 50.2 & 32.0 & 16.6 & 34.1 & 41.1\\
			MixNet-M \cite{tan2020mixconv} & 360 & 5.0 & 77.0 & 31.3& 51.7 & 32.4& 17.0 & 35.0 & 41.9   \\
			FairNAS-A$^{\dagger}$ & 392 & 5.9 & \textbf{77.5} & \textbf{32.4} & \textbf{52.4} & \textbf{33.9} & \textbf{17.2} & \textbf{36.3} & \textbf{43.2}\\
			FairNAS-B$^{\dagger}$ & 349 & 5.7 & 77.2 & 31.7 & 51.5 & 33.0 & 17.0 & 35.2 & 42.5\\
			FairNAS-C$^{\dagger}$ &325 & 5.6& 76.7& 31.2 & 50.8 & 32.7 & 16.3 & 34.4 & 42.3\\
			\hline 
			\noalign{\smallskip}
		\end{tabular}
		\caption{Object detection on COCO with various drop-in backbones. $^{\dagger}$: w/ SE and Swish}
		\label{table:fairnas-coco-retina}
\end{table}
\setlength{\tabcolsep}{1.4pt}

\subsection{Transferability on Semantic Segmentation}

We further evaluate FairNAS models as a feature extractor with DeepLabv3+\cite{deeplabv3plus2018} on the mobile semantic segmentation task, which confirms FairNAS backbones are competitive. All models are first pre-trained on COCO dataset \cite{lin2014coco}, then coarsely trained on VOC2012 \cite{pascal-voc-2012} extra annotated images and fine-tuned on VOC2012 fine annotated images. The results are given in Table~\ref{tab:segment-voc} where  the Atrous Spatial Pyramid Pooling (ASPP) module and multi-scale contextual information are not used. We also don't flip inputs left or right during test.

\begin{table}[ht]
	\begin{center}
		\begin{small}
				\begin{tabular}{|l|l|c|Hl|l|l|}
					\hline 
					Network & OS & ASPP & MF & Params& $\times+$& mIOU  \\
					\hline 
					MobileNetV1 \cite{howard2017mobilenets} & 16 & \cmark & &11.15M & 14.25B &75.29 \\
					MobileNetV2 \cite{sandler2018mobilenetv2} & 16 & \xmark & \xmark & 4.52M & 2.75B & 75.32\\
					FairNAS-A	& 16	& \xmark& & 3.26M & 3.98B & \textbf{78.54} \\
					FairNAS-B & 16 & \xmark& & 3.11M & 3.74B & 77.10 \\
					FairNAS-C & 16 & \xmark & &  3.01M & 3.60B & 77.64 \\
					\hline
				\end{tabular}
		\end{small}
	\end{center}
	\caption{Semantic Segmentation on VOC 2012. OS: Output Stride}
	\label{tab:segment-voc}
\end{table}

\section{Ablation Study}

\subsection{Model Ranking Capacity} 

As stated, the most important role of the supernet in the two-stage methods is to score models' relative performance, i.e. model ranking.

For supernet training, we set up three control groups that meet \emph{Expectation Fairness}  as our baselines. 
\begin{enumerate*} 
	\item \textbf{EF} $lr$, uniformly sampling one path and train $k$ times, followed by parameter update. 
	\item \textbf{EF} $1/6 lr$: same as the first one except that the learning rate is scaled by $\frac{1}{k}$. In practice, we set $k=6$ to make it comparable to FairNAS. 
	\item \textbf{SPOS} : an reimplementation of Single-Path One-Shot \cite{guo2019single}. 
\end{enumerate*}
Other hyperparameters are kept the same. Note a), c) and FairNAS all use the same learning rate $lr$. 

We run the search pipeline for 200 epochs with a population size of 64, sampling 12,800 models in total. It takes only 2 GPU days due to accelerated evaluation. Due to high training cost, we sampled 13 models at approximately equal distances on the Pareto front and trained them from scratch to get the ranking, which is shown to the right of Figure~\ref{fig:fairness-hist}. We observe that the FairNAS supernet gives a highly relevant ranking while Single-Path One-Shot \cite{guo2019single} doesn't. The training process of sampled models is plotted in Figure \ref{fig:standalone_val} (see supplementary).






We further adopt Kendall Tau \cite{kendall1938new} for the ranking analysis following a recent work \cite{sciuto2019evaluating} that evaluates NAS approaches. A method based on incomplete training reaches an average $\tau$ of 0.474 \cite{zheng2019multi}.
Instead, we hit a new high record of the Kendall rank correlation coefficient $\tau=0.9487$. We show our ranking comparison with baseline groups in Table~\ref{tab:kendall-tau}. In general,  \textbf{methods with EF have a better ranking than those without EF, while SF is the best of all}, which discloses the relevance of fairness to ranking in one-shot approaches.

\setlength{\tabcolsep}{4pt}
\begin{table}
	\centering
		\begin{small}
			\begin{tabular}{|l|*{3}{c|}} 			
				\hline
				Methods & Fairness & $\tau_{a}$  &$\tau_{N}$ \\
				\hline
				One-Shot \cite{bender2018understanding}$^\dagger$ & None & 0.1245 & 0.0934\\
				Uniform ($k=6$, baseline) & EF & 0.4871 & 0.3651 \\
				Uniform ($k=1$, $1/6 lr$) & EF & 0.4871 & 0.4072 \\
				SPOS \cite{guo2019single} ($k=1$)$^\dagger$ & EF & 0.6153 &0.5681 \\
				FairNAS$^\ddagger$ & SF & \textbf{0.9487} & \textbf{0.7412} \\
				\hline
				\noalign{\smallskip}
			\end{tabular}
		\end{small}
	\caption{Ranking ability of methods satisfying Expected Strictness vs. Strict Strictness in NAS-Bench-201 ($\tau_N$) and in search space (a) ($\tau_a$). For the latter, 13 models are fully trained on ImageNet to obtain their ground-truth ranking order.  $^\dagger$: Reimplemented. $^\ddagger$: With or without recalculating batch normalization, $\tau$ holds the same. For EF methods, $k=6$ iterations are performed at each training step}
	\label{tab:kendall-tau}
\end{table}

\subsection{Comparisons of Searching Algorithms}

For the second-stage, we adopt multi-objective optimization where three objectives are considered: accuracies, multiply-adds, and the number of parameters. Specifically, we apply MoreMNAS with a minor modification in which PPO \cite{schulman2017proximal} is utilized instead of REINFORCE \cite{sutton2018reinforcement}. 

\begin{figure}[ht]
\centering
\includegraphics[scale=0.3,width=.46\columnwidth]{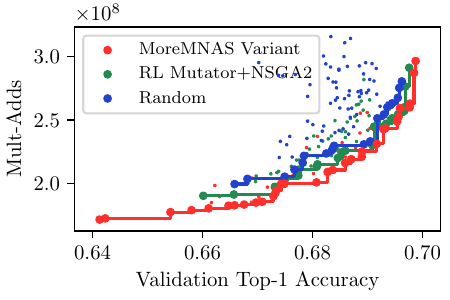}
\includegraphics[scale=0.3,width=.46\columnwidth]{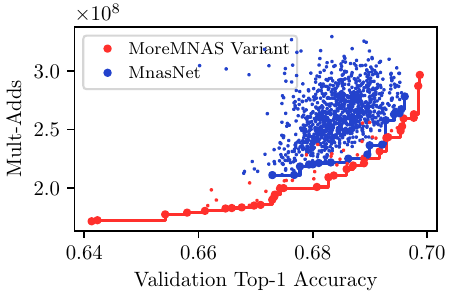}
\caption{Pareto front (last generation elitists $P_{G+1}$) of the MoreMNAS variant (adopted) compared with \textbf{Left:} NSGA2 (EA-like baseline) with RL mutator and random search (random baseline), 
\textbf{Right:} MnasNet (RL baseline). Each samples 1,088 models.} 
\label{fig:ablation_model}
\vskip -0.1in
\end{figure}

We construct several comparison groups that cover the main searching algorithms: 
\begin{enumerate*}
\item \textbf{EA}: NSGA-II with reinforced mutation, 
\item \textbf{random search},
\item \textbf{Multi-objective RL}: MnasNet which uses PPO with a mixed multi-objective reward \cite{tan2018mnasnet}. 
\end{enumerate*}
The results are shown in Figure~\ref{fig:ablation_model}, control groups generally align within our Pareto front and are constricted within a narrow range, affirming an excellent advantage in the MoreMNAS variant. 

\subsection{Component Contribution Analysis} 

Being a two-stage method, which stage contributes more to the final performance of the architecture?
The experiment in NAS-Bench-201 has answered this question. To be complete,  we further compare various supernet training strategies while fixing the second stage using the ImageNet dataset.  Considering it's not affordable to train the entire models from Pareto front, we impose an explicit constraint of maximum 400M FLOPS. 
The best models from One-Shot [2] (no EF) and SPOS (EF) [7] reach  $74.0\%$ and $74.6\%$ top-1 accuracies, indicating
 that our result ($75.3\%$) benefits mainly from the ranking capacity of the supernet in the first stage.






\section{Discussions}
\subsection{Why Does Single-Path Training Work?}

\begin{figure}[ht]
\centering
\includegraphics[scale=0.5]{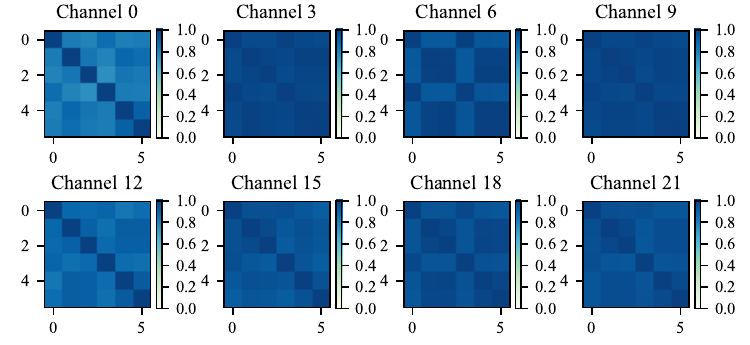}
\caption{Cross-block channel-wise cosine similarity matrix on feature maps of 6 choice blocks in Layer 1. We observe that each choice block learns very similar features on the same channel}
\label{fig:bird_channelwise_cosine_distance_selected}
\end{figure}

Our supernet generates a relatively small range of one-shot accuracies, from which we postulate that choice blocks be quite alike in terms of capacity. In fact, given an input of a chickadee image, the choice blocks of the first layer yield similar feature maps on the same channel, as shown in Figure~\ref{fig:fair_activation}.  But how much do they resemble each other? We involve the \emph{cosine similarity} \cite{nguyen2010cosine} to measure the distance among various feature vectors. It ranges from -1 (opposite)  to 1 (identical), where 0 indicates no correlation. In Figure~\ref{fig:bird_channelwise_cosine_distance_selected}, each $6\times6$ symmetric matrix shows the cross-block distances per channel, they are very similar (above 0.9). 


In summary, \emph{the channel-wise feature maps generated by our supernet come with high similarities}. We conclude that this important characteristic significantly stabilizes the whole training process. For layer $l+1$, its input is randomly from choice blocks in the previous layer $l$. As different choices have highly similar channel-aligned features, the random sampling constructs a mechanism mimicking  \textbf{feature augmentation}, which boosts the supernet training. 

\subsection{Fairness Closes Supernet Accuracy gap} 
As discussed in Section \ref{sec:review-bias}, previous one-shot methods \cite{brock2017smash,bender2018understanding} have a large accuracy gap between the one-shot and stand-alone models. We define it as \emph{supernet accuracy gap}, $\lambda=|\delta_{oneshot}-\delta_{standalone}|$,
where $\delta_{oneshot}$ is the accuracy range of one-shot models, and $\delta_{standalone}$ for stand-alone models. Ideally, $\delta_{oneshot}$ can be obtained by evaluating all paths from the supernet but not affordable since the search space is enormous. Instead, we can approximate $\delta_{oneshot}$ by covering a wide range of models. We randomly sample 1,000 models from our supernet, then we evaluate these models directly on the ImageNet validation set. Their top-1 accuracies (see Figure~\ref{fig:fairness-hist}) range from 0.666 to 0.696, which leads to $\delta_{oneshot}=0.03$, hence it reduces $\lambda$ as well. 


\section{Conclusion}
In this work, we scrutinize the weight-sharing neural architecture search with a fairness perspective. Observing that unfairness inevitably incurs a severely biased evaluation of one-shot model performance, we propose two degrees of fairness enhancement, where \emph{Strict Fairness} (SF) works best. Our supernet trained under SF then acts as a performance evaluator. 
In principle, the fair supernet can be incorporated in any search pipeline that requires an evaluator.  To demonstrate its effectiveness, we adopt a multi-objective evolutionary backend. After searching proxylessly on ImageNet for 12 GPU days, we harvest three state-of-the-art models of different magnitudes nearby \textit{Pareto Optimality}. Future works remain as to study fairness under heterogenous search spaces and to improve the evaluation performance of the supernet.  

\clearpage
{
\small
\bibliographystyle{ieee_fullname}
\bibliography{egbib}

\begin{thebibliography}{10}\itemsep=-1pt

\bibitem{bender2018understanding}
Gabriel Bender, Pieter-Jan Kindermans, Barret Zoph, Vijay Vasudevan, and Quoc
  Le.
\newblock {Understanding and Simplifying One-Shot Architecture Search}.
\newblock In {\em International Conference on Machine Learning}, pages
  549--558, 2018.

\bibitem{brock2017smash}
Andrew Brock, Theodore Lim, James~M Ritchie, and Nick Weston.
\newblock {SMASH: One-Shot Model Architecture Search through HyperNetworks}.
\newblock In {\em International Conference on Learning Representations}, 2018.

\bibitem{cai2019once}
Han Cai, Chuang Gan, and Song Han.
\newblock {Once for All: Train One Network and Specialize it for Efficient
  Deployment}.
\newblock In {\em International Conference on Learning Representations}, 2020.

\bibitem{cai2018proxylessnas}
Han Cai, Ligeng Zhu, and Song Han.
\newblock {Proxyless{NAS}: Direct Neural Architecture Search on Target Task and
  Hardware}.
\newblock In {\em International Conference on Learning Representations}, 2019.

\bibitem{chen2019mmdetection}
Kai Chen, Jiaqi Wang, Jiangmiao Pang, Yuhang Cao, Yu Xiong, Xiaoxiao Li,
  Shuyang Sun, Wansen Feng, Ziwei Liu, Jiarui Xu, et~al.
\newblock {MMDetection: Open mmlab detection toolbox and benchmark}.
\newblock {\em arXiv preprint arXiv:1906.07155}, 2019.

\bibitem{deeplabv3plus2018}
Liang-Chieh Chen, Yukun Zhu, George Papandreou, Florian Schroff, and Hartwig
  Adam.
\newblock {Encoder-Decoder with Atrous Separable Convolution for Semantic Image
  Segmentation}.
\newblock In {\em European Conference on Computer Vision}, 2018.

\bibitem{chu2019multi}
Xiangxiang Chu, Bo Zhang, and Ruijun Xu.
\newblock {Multi-Objective Reinforced Evolution in Mobile Neural Architecture
  Search}.
\newblock In {\em European Conference on Computer Vision Workshop}, 2020.

\bibitem{cubuk2018autoaugment}
Ekin~D Cubuk, Barret Zoph, Dandelion Mane, Vijay Vasudevan, and Quoc~V Le.
\newblock {AutoAugment: Learning Augmentation Policies from Data}.
\newblock {\em Proceedings of the IEEE Conference on Computer Vision and
  Pattern Recognition}, 2019.

\bibitem{deb2002fast}
Kalyanmoy Deb, Amrit Pratap, Sameer Agarwal, and TAMT Meyarivan.
\newblock {A Fast and Elitist Multi-objective Genetic Algorithm: NSGA-II}.
\newblock {\em IEEE Transactions on Evolutionary Computation}, 6(2):182--197,
  2002.

\bibitem{devries2017improved}
Terrance DeVries and Graham~W Taylor.
\newblock {Improved Regularization of Convolutional Neural Networks with
  Cutout}.
\newblock {\em arXiv preprint. arXiv:1708.04552}, 2017.

\bibitem{dong2019one}
Xuanyi Dong and Yi Yang.
\newblock {One-Shot Neural Architecture Search via Self-Evaluated Template
  Network}.
\newblock In {\em Proceedings of the IEEE International Conference on Computer
  Vision}, pages 3681--3690, 2019.

\bibitem{dong2019search}
Xuanyi Dong and Yi Yang.
\newblock {Searching for A Robust Neural Architecture in Four GPU Hours}.
\newblock In {\em Proceedings of the IEEE Conference on Computer Vision and
  Pattern Recognition}, pages 1761--1770, 2019.

\bibitem{dong2019bench}
Xuanyi Dong and Yi Yang.
\newblock {NAS-Bench-201: Extending the Scope of Reproducible Neural
  Architecture Search}.
\newblock In {\em International Conference on Learning Representations}, 2020.

\bibitem{pascal-voc-2012}
M. Everingham, L. Van~Gool, C.~K.~I. Williams, J. Winn, and A. Zisserman.
\newblock The {PASCAL} {V}isual {O}bject {C}lasses {C}hallenge 2012 {(VOC2012)}
  {R}esults.
\newblock
  http://www.pascal-network.org/challenges/VOC/voc2012/workshop/index.html,
  2012.

\bibitem{guo2019single}
Zichao Guo, Xiangyu Zhang, Haoyuan Mu, Wen Heng, Zechun Liu, Yichen Wei, and
  Jian Sun.
\newblock {Single Path One-Shot Neural Architecture Search with Uniform
  Sampling}.
\newblock {\em European Conference on Computer Vision}, 2020.

\bibitem{han2020ghostnet}
Kai Han, Yunhe Wang, Qi Tian, Jianyuan Guo, Chunjing Xu, and Chang Xu.
\newblock {GhostNet: More Features from Cheap Operations}.
\newblock In {\em Proceedings of the IEEE Conference on Computer Vision and
  Pattern Recognition}, 2020.

\bibitem{hochreiter1997long}
Sepp Hochreiter and J{\"u}rgen Schmidhuber.
\newblock {Long Short-Term Memory}.
\newblock {\em Neural computation}, 9(8):1735--1780, 1997.

\bibitem{howard2019searching}
Andrew Howard, Mark Sandler, Grace Chu, Liang-Chieh Chen, Bo Chen, Mingxing
  Tan, Weijun Wang, Yukun Zhu, Ruoming Pang, Vijay Vasudevan, et~al.
\newblock {Searching for MobileNetV3}.
\newblock In {\em International Conference on Computer Vision}, 2019.

\bibitem{howard2017mobilenets}
Andrew~G Howard, Menglong Zhu, Bo Chen, Dmitry Kalenichenko, Weijun Wang,
  Tobias Weyand, Marco Andreetto, and Hartwig Adam.
\newblock {MobileNets: Efficient Convolutional Neural Networks for Mobile
  Vision Applications}.
\newblock {\em arXiv preprint arXiv:1704.04861}, 2017.

\bibitem{hu2018squeeze}
Jie Hu, Li Shen, and Gang Sun.
\newblock {Squeeze-and-Excitation Networks}.
\newblock In {\em Proceedings of the IEEE Conference on Computer Vision and
  Pattern Recognition}, pages 7132--7141, 2018.

\bibitem{huang2018gpipe}
Yanping Huang, Yonglong Cheng, Dehao Chen, HyoukJoong Lee, Jiquan Ngiam, Quoc~V
  Le, and Zhifeng Chen.
\newblock {GPipe: Efficient Training of Giant Neural Networks using Pipeline
  Parallelism}.
\newblock {\em Neural Information Processing Systems}, 2019.

\bibitem{kendall1938new}
Maurice~G Kendall.
\newblock {A New Measure of Rank Correlation}.
\newblock {\em Biometrika}, 30(1/2):81--93, 1938.

\bibitem{kornblith2019better}
Simon Kornblith, Jonathon Shlens, and Quoc~V Le.
\newblock {Do Better Imagenet Models Transfer Better?}
\newblock In {\em Proceedings of the IEEE Conference on Computer Vision and
  Pattern Recognition}, pages 2661--2671, 2019.

\bibitem{krizhevsky2009learning}
Alex Krizhevsky, Geoffrey Hinton, et~al.
\newblock {Learning Multiple Layers of Features from Tiny Images}.
\newblock Technical report, Citeseer, 2009.

\bibitem{li2019random}
Liam Li and Ameet Talwalkar.
\newblock {Random Search and Reproducibility for Neural Architecture Search}.
\newblock {\em Conference on Uncertainty in Artificial Intelligence}, 2019.

\bibitem{liang2019darts+}
Hanwen Liang, Shifeng Zhang, Jiacheng Sun, Xingqiu He, Weiran Huang, Kechen
  Zhuang, and Zhenguo Li.
\newblock {DARTS+: Improved Differentiable Architecture Search with Early
  Stopping}.
\newblock {\em arXiv preprint arXiv:1909.06035}, 2019.

\bibitem{lin2017focal}
Tsung-Yi Lin, Priya Goyal, Ross Girshick, Kaiming He, and Piotr Dollár.
\newblock {Focal Loss for Dense Object Detection}.
\newblock In {\em International Conference on Computer Vision}, 2017.

\bibitem{lin2014coco}
Tsung-Yi Lin, Michael Maire, Serge~J. Belongie, Lubomir~D. Bourdev, Ross~B.
  Girshick, James Hays, Pietro Perona, Deva Ramanan, Piotr Dollár, and
  C.~Lawrence Zitnick.
\newblock {Microsoft COCO: Common Objects in Context}.
\newblock In {\em European Conference on Computer Vision}, 2014.

\bibitem{liu2018darts}
Hanxiao Liu, Karen Simonyan, and Yiming Yang.
\newblock {DARTS: Differentiable Architecture Search}.
\newblock In {\em International Conference on Learning Representations}, 2019.

\bibitem{loshchilov2016sgdr}
Ilya Loshchilov and Frank Hutter.
\newblock {SGDR: Stochastic Gradient Descent with Warm Restarts}.
\newblock In {\em International Conference on Learning Representations}, 2017.

\bibitem{lu2018nsga}
Zhichao Lu, Ian Whalen, Vishnu Boddeti, Yashesh Dhebar, Kalyanmoy Deb, Erik
  Goodman, and Wolfgang Banzhaf.
\newblock {NSGA-NET: A Multi-Objective Genetic Algorithm for Neural
  Architecture Search}.
\newblock In {\em Proceedings of the Genetic and Evolutionary Computation
  Conference}, pages 419--427, 2019.

\bibitem{luo2018neural}
Renqian Luo, Fei Tian, Tao Qin, Enhong Chen, and Tie-Yan Liu.
\newblock {Neural Architecture Optimization}.
\newblock In {\em Advances in Neural Information Processing Systems}, pages
  7816--7827, 2018.

\bibitem{Mei2020AtomNAS:}
Jieru Mei, Yingwei Li, Xiaochen Lian, Xiaojie Jin, Linjie Yang, Alan Yuille,
  and Jianchao Yang.
\newblock {AtomNAS: Fine-Grained End-to-End Neural Architecture Search}.
\newblock In {\em International Conference on Learning Representations}, 2020.

\bibitem{nguyen2010cosine}
Hieu~V Nguyen and Li Bai.
\newblock {Cosine Similarity Metric Learning for Face Verification}.
\newblock In {\em Asian Conference on Computer Vision}, pages 709--720.
  Springer, 2010.

\bibitem{pham2018efficient}
Hieu Pham, Melody~Y Guan, Barret Zoph, Quoc~V Le, and Jeff Dean.
\newblock {Efficient Neural Architecture Search via Parameter Sharing}.
\newblock In {\em International Conference on Machine Learning}, 2018.

\bibitem{ramachandran2017searching}
Prajit Ramachandran, Barret Zoph, and Quoc~V. Le.
\newblock {Searching for Activation Functions}.
\newblock {\em arXiv preprint arXiv:1710.05941}, 2017.

\bibitem{real2018regularized}
Esteban Real, Alok Aggarwal, Yanping Huang, and Quoc~V Le.
\newblock {Regularized Evolution for Image Classifier Architecture Search}.
\newblock {\em International Conference on Machine Learning, AutoML Workshop},
  2018.

\bibitem{russakovsky2015imagenet}
Olga Russakovsky, Jia Deng, Hao Su, Jonathan Krause, Sanjeev Satheesh, Sean Ma,
  Zhiheng Huang, Andrej Karpathy, Aditya Khosla, Michael Bernstein, et~al.
\newblock {ImageNet Large Scale Visual Recognition Challenge}.
\newblock {\em International Journal of Computer Vision}, 115(3):211--252,
  2015.

\bibitem{sandler2018mobilenetv2}
Mark Sandler, Andrew Howard, Menglong Zhu, Andrey Zhmoginov, and Liang-Chieh
  Chen.
\newblock {MobileNetV2: Inverted Residuals and Linear Bottlenecks}.
\newblock In {\em Proceedings of the IEEE Conference on Computer Vision and
  Pattern Recognition}, pages 4510--4520, 2018.

\bibitem{schulman2017proximal}
John Schulman, Filip Wolski, Prafulla Dhariwal, Alec Radford, and Oleg Klimov.
\newblock {Proximal Policy Optimization Algorithms}.
\newblock {\em arXiv preprint. arXiv:1707.06347}, 2017.

\bibitem{sciuto2019evaluating}
Christian Sciuto, Kaicheng Yu, Martin Jaggi, Claudiu Musat, and Mathieu
  Salzmann.
\newblock {Evaluating the Search Phase of Neural Architecture Search}.
\newblock {\em International Conference on Learning Representations}, 2020.

\bibitem{stamoulis2019single}
Dimitrios Stamoulis, Ruizhou Ding, Di Wang, Dimitrios Lymberopoulos, Bodhi
  Priyantha, Jie Liu, and Diana Marculescu.
\newblock {Single-Path NAS: Designing Hardware-Efficient ConvNets in less than
  4 Hours}.
\newblock In {\em European Conference on Machine Learning and Principles and
  Practice of Knowledge Discovery in Databases}, 2019.

\bibitem{sutskever2013importance}
Ilya Sutskever, James Martens, George Dahl, and Geoffrey Hinton.
\newblock {On the Importance of Initialization and Momentum in Deep Learning}.
\newblock In {\em International Conference on Machine Learning}, pages
  1139--1147, 2013.

\bibitem{sutton2018reinforcement}
Richard~S Sutton and Andrew~G Barto.
\newblock {\em {Reinforcement Learning: An Introduction}}.
\newblock MIT press, 2018.

\bibitem{tan2018mnasnet}
Mingxing Tan, Bo Chen, Ruoming Pang, Vijay Vasudevan, and Quoc~V Le.
\newblock {MnasNet: Platform-Aware Neural Architecture Search for Mobile}.
\newblock In {\em Proceedings of the IEEE Conference on Computer Vision and
  Pattern Recognition}, 2019.

\bibitem{tan2019efficientnet}
Mingxing Tan and Quoc~V Le.
\newblock {EfficientNet: Rethinking Model Scaling for Convolutional Neural
  Networks}.
\newblock In {\em International Conference on Machine Learning}, 2019.

\bibitem{tan2020mixconv}
Mingxing Tan and Quoc~V. Le.
\newblock {MixConv: Mixed Depthwise Convolutional Kernels}.
\newblock {\em The British Machine Vision Conference}, 2019.

\bibitem{tweddle2012james}
Ian Tweddle.
\newblock {\em {James Stirling’s Methodus Differentialis: An Annotated
  Translation of Stirling’s Text}}.
\newblock Springer Science \& Business Media, 2012.

\bibitem{wu2018fbnet}
Bichen Wu, Xiaoliang Dai, Peizhao Zhang, Yanghan Wang, Fei Sun, Yiming Wu,
  Yuandong Tian, Peter Vajda, Yangqing Jia, and Kurt Keutzer.
\newblock {FBNet: Hardware-Aware Efficient ConvNet Design via Differentiable
  Neural Architecture Search}.
\newblock In {\em Proceedings of the IEEE Conference on Computer Vision and
  Pattern Recognition}, 2019.

\bibitem{Xu2020PC-DARTS:}
Yuhui Xu, Lingxi Xie, Xiaopeng Zhang, Xin Chen, Guo-Jun Qi, Qi Tian, and
  Hongkai Xiong.
\newblock {PC-DARTS: Partial Channel Connections for Memory-Efficient
  Architecture Search}.
\newblock In {\em International Conference on Learning Representations}, 2020.

\bibitem{yu2020bignas}
Jiahui Yu, Pengchong Jin, Hanxiao Liu, Gabriel Bender, Pieter-Jan Kindermans,
  Mingxing Tan, Thomas Huang, Xiaodan Song, Ruoming Pang, and Quoc Le.
\newblock {BigNAS: Scaling up Neural Architecture Search with Big Single-stage
  Models}.
\newblock In {\em European Conference on Computer Vision}, pages 702--717.
  Springer, 2020.

\bibitem{zela2019understanding}
Arber Zela, Thomas Elsken, Tonmoy Saikia, Yassine Marrakchi, Thomas Brox, and
  Frank Hutter.
\newblock {Understanding and Robustifying Differentiable Architecture Search}.
\newblock {\em International Conference on Learning Representations}, 2020.

\bibitem{zhang2018mixup}
Hongyi Zhang, Moustapha Cisse, Yann~N. Dauphin, and David Lopez-Paz.
\newblock {mixup: Beyond Empirical Risk Minimization}.
\newblock In {\em International Conference on Learning Representations}, 2018.

\bibitem{zheng2019multi}
Xiawu Zheng, Rongrong Ji, Lang Tang, Baochang Zhang, Jianzhuang Liu, and Qi
  Tian.
\newblock {Multinomial Distribution Learning for Effective Neural Architecture
  Search}.
\newblock In {\em International Conference on Computer Vision}, 2019.

\bibitem{zoph2016neural}
Barret Zoph and Quoc~V Le.
\newblock {Neural Architecture Search with Reinforcement Learning}.
\newblock In {\em International Conference on Learning Representations}, 2017.

\bibitem{zoph2017learning}
Barret Zoph, Vijay Vasudevan, Jonathon Shlens, and Quoc~V Le.
\newblock {Learning Transferable Architectures for Scalable Image Recognition}.
\newblock In {\em Proceedings of the IEEE Conference on Computer Vision and
  Pattern Recognition}, volume~2, 2018.

\end{thebibliography}
}

\newpage{}
\appendix
\section{More Discussion of Algorithms}\label{app:alg}
\subsection{Supernet Training}
Figure~\ref{fig:one-shot-strategy} details the supernet training stage of our approach. In fact, it's inherently efficient regarding GPU utilization. Even on powerful machines such as Tesla V100, it can make full use of GPU without special optimization. As most of the existing deep learning frameworks allow paralleled execution between data generation and gradient calculation, our algorithm can exploit this parallelism to the extreme since a mini-batch of data is reused by $m$ times of backpropagation. The GPUs are always busy because the data is ready whenever required, which shortens the training time. Moreover, our method works in a single-path way, which is memory friendly.

\textbf{Irregular Search Spaces.} Note that SF in the paper can be easily extended by a preprocessing function in case of irregular search spaces (i.e. the number of operations are not the same for each layer). We only need to make a minor modification of Algorithm~\ref{alg:fair_sample}. Say the $l$-th layer has $m_l$ choices.  Suppose $M =  \max (m_l)$, we randomly choose $M-m_l$ extra operations from $m_l$ choices and regard these extra options as different ones from the original search space. Therefore, the input condition of Algorithm~\ref{alg:fair_sample} still hold and we can use it directly. This procedure can be regarded as an approximated SF.  However, perfect SF for irregular cases remains as our future work.

\subsection{Evolutionary Searching Pipeline}\label{sec:pipeline}
\begin{figure}[ht]
	\centering
	\includegraphics[width=.45\textwidth]{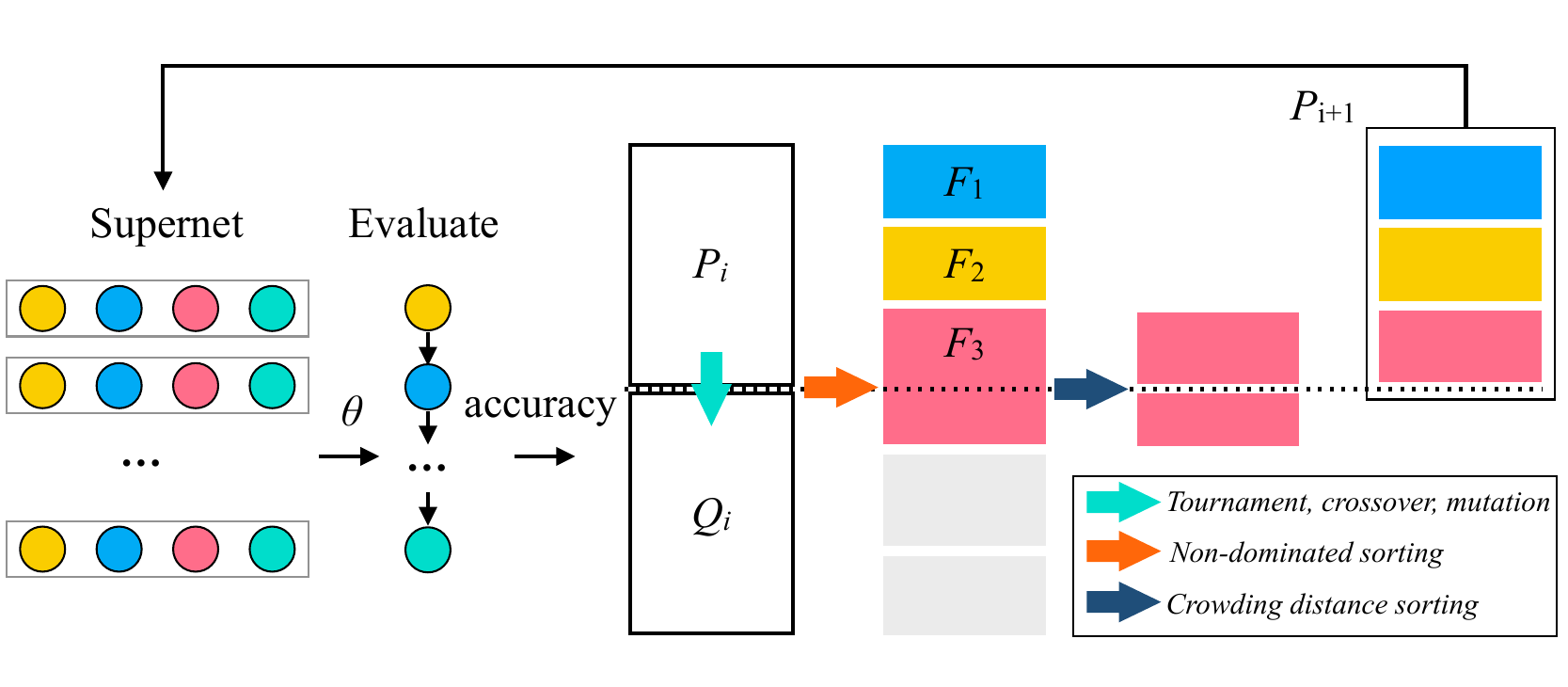}
	\caption{Evolutionary searching with the supernet trained with \emph{strict fairness}. In each generation, candidate models in the current population inherit weights from the supernet for evaluation. Their estimated accuracies are fed into the searching pipeline as one of the objectives. The evolution loops till Pareto optimality.}
	\label{fig:fairnas}
\end{figure}

With our supernet fairly trained as a model evaluator, we adopt an evolutionary-based algorithm for searching, detailed in Algorithm~\ref{alg:nas_pipeline} (main text) and Figure~\ref{fig:fairnas}. Generally, it is built on the ground of MoreMNAS \cite{chu2019multi} by replacing its incomplete-training evaluator with our fairly trained supernet. FairNAS supernet exhibits tremendous speed-up in terms of GPU days by two orders of magnitudes. We also use Proximal Policy Optimization as the default reinforcement algorithm \cite{schulman2017proximal}.

\section{A Fairness Taxonomy}
We compare current weight-sharing NAS methods based on the defined fairness in Table \ref{tab:comparison-nas}. SPOS \cite{guo2019single} satisfies Expectation Fairness, while FairNAS meets Strict Fairness.

\begin{table}
\setlength{\tabcolsep}{2pt}
	\begin{center}
	  \begin{footnotesize}
		\begin{tabular}{*{2}{l}H*{4}{l}} 
			\toprule
			NAS Methods & Type  & $M$  & $C_t$ & $C_s$ &EF & SF \\
			\midrule
			 SMASH \cite{brock2017smash} & Hypernet &  SN & - & -  &  \xmark & \xmark \\
			 One-Shot \cite{bender2018understanding}   & Supernet & SN & 4$^\ddagger$ & 3.3 &\xmark & \xmark \\ %
			 DARTS \cite{liu2018darts}  & Gradient-based & SN & 0.5$^\dagger$ & 0 & \xmark & \xmark  \\	 %
			 FBNet  \cite{wu2018fbnet}&Gradient-based & \textbf{SP} & 9& 0 &\xmark&\xmark \\ %
			 ProxylessNAS \cite{cai2018proxylessnas}  & Gradient-based/RL & TP  & 8.3 & 0 &   \xmark  & \xmark\\ %
			 SPOS  \cite{guo2019single} &  Supernet+EA &  \textbf{SP}  & 12 & $<$1 &\xmark& \xmark \\ %
			 Single-Path NAS  \cite{stamoulis2019single}  & Gradient-based & \textbf{SP} &  1.25$^\ddagger$ & 0 &\cmark&\xmark \\ 
			 FairNAS (Ours) & Fair Supernet+EA & \textbf{SP} &   10 & 2 & \cmark & \cmark\\
			\bottomrule
		\end{tabular}
		\end{footnotesize}
	\end{center}
	\caption{Comparison of state-of-the-art weight-sharing NAS methods as per cost and fairness basis. $C_t, C_s$: train and search cost measured in GPU days. EF: Expectation Fairness, SF: Strict Fairness. $^\dagger$: searched on CIFAR-10, $^\ddagger$: TPU} 
	\label{tab:comparison-nas}
\end{table}

\section{Experiment Details}

\textbf{Dataset.}
The supernet experiments are performed on ImageNet \cite{russakovsky2015imagenet} and we randomly select 50,000 images from the training set as our validation set (50 samples from each class). The remaining training set is used as our training set, while the original validation set is taken as the test set to measure the final performance of each model. 

\subsection{Architectures of Searched Models}

The searched FairNAS-A, B and C models are illustrated in Figure~\ref{fig:fairnas-a}.

\begin{figure*}[ht]
	\begin{center}
		\subfigure{
			\centerline{
				\begin{tikzpicture}[thick,scale=0.5, every node/.style={scale=0.5},node distance=1.3cm]
				\node (s1) [cell_r,label={[rotate=90,xshift=0.1cm,yshift=0.5cm]right:3*224*224}] {Conv\_K3};
				\node (s2) [cell_r_b, yshift=-1.3cm,xshift=-1.3cm, right of=s1,label={[rotate=90,xshift=0.1cm,yshift=0.3cm]right:32*112*112}] {MBE1\_K3};
				\node (a) [cell_r_p, e3, yshift=-1.3cm,xshift=-1.3cm, right of=s2,label={[rotate=90,xshift=0.1cm,yshift=0.3cm]right:16*112*112}] {MBE3\_K7}; 
				\node (b) [cell_r_b,e3 , yshift=-1.3cm,xshift=-1.3cm, right of=a,label={[rotate=90,xshift=0.1cm,yshift=0.3cm]right:32*56*56}] {MBE3\_K3};
				\node (c) [cell_r_p,e3 , yshift=-1.3cm,xshift=-1.3cm, right of=b,label={[rotate=90,xshift=0.1cm,yshift=0.3cm]right:32*56*56}] {MBE3\_K7}; 
				\node (d) [cell_r_b,e6 , yshift=-1.3cm,xshift=-1.3cm, right of=c,label={[rotate=90,xshift=-0.3cm,yshift=0.3cm]right:40*28*28}] {MBE6\_K3};
				\node (e) [cell_r_p,e6 , yshift=-1.3cm,xshift=-1.3cm, right of=d,label={[rotate=90,xshift=-0.3cm,yshift=0.3cm]right:40*28*28}] {MBE6\_K7};
				\node (f) [cell_r_b, e3 , yshift=-1.3cm,xshift=-1.3cm, right of=e,label={[rotate=90,xshift=0.1cm,yshift=0.3cm]right:40*28*28}] {MBE3\_K3};
				\node (g) [cell_r_b, e3 , yshift=-1.3cm,xshift=-1.3cm, right of=f,label={[rotate=90,xshift=0.1cm,yshift=0.3cm]right:40*28*28}] {MBE3\_K3};
				\node (h) [cell_r_p, e6 , yshift=-1.3cm,xshift=-1.3cm, right of=g,label={[rotate=90,xshift=-0.3cm,yshift=0.3cm]right:80*14*14}] {MBE6\_K7};
				\node (i) [cell_r_p, e6 , yshift=-1.3cm,xshift=-1.3cm, right of=h,label={[rotate=90,xshift=-0.3cm,yshift=0.5cm]right:80*14*14}] {MBE6\_K7};
				\node (j) [cell_r_g, e3 , yshift=-1.3cm,xshift=-1.3cm, right of=i,label={[rotate=90,xshift=0.1cm,yshift=0.3cm]right:80*14*14}] {MBE3\_K5};
				\node (k) [cell_r_b, e6 , yshift=-1.3cm,xshift=-1.3cm, right of=j,label={[rotate=90,xshift=-0.3cm,yshift=0.5cm]right:80*14*14}] {MBE6\_K3};
				\node (l) [cell_r_g, e3 , yshift=-1.3cm,xshift=-1.3cm, right of=k,label={[rotate=90,xshift=0.1cm,yshift=0.5cm]right:96*14*14}] {MBE3\_K5};
				\node (m) [cell_r_g, e3 , yshift=-1.3cm,xshift=-1.3cm, right of=l,label={[rotate=90,xshift=0.1cm,yshift=0.5cm]right:96*14*14}] {MBE3\_K5}; 
				\node (n) [cell_r_b, e6 , yshift=-1.3cm,xshift=-1.3cm, right of=m,label={[rotate=90,xshift=-0.3cm,yshift=0.5cm]right:96*14*14}] {MBE6\_K3};
				\node (o) [cell_r_b, e6 , yshift=-1.3cm,xshift=-1.3cm, right of=n,label={[rotate=90,xshift=-0.3cm,yshift=0.5cm]right:96*14*14}] {MBE6\_K3}; 
				\node (p) [cell_r_p, e6 , yshift=-1.3cm,xshift=-1.3cm, right of=o,label={[rotate=90,xshift=-0.3cm,yshift=0.3cm]right:192*7*7}] {MBE6\_K7};
				\node (q) [cell_r_b, e6 , yshift=-1.3cm,xshift=-1.3cm, right of=p,label={[rotate=90,xshift=-0.3cm,yshift=0.3cm]right:192*7*7}] {MBE6\_K3};
				\node (r) [cell_r_p, e6 , yshift=-1.3cm,xshift=-1.3cm, right of=q,label={[rotate=90,xshift=-0.3cm,yshift=0.3cm]right:192*7*7}] {MBE6\_K7};
				\node (s) [cell_r_g, e6 , yshift=-1.3cm,xshift=-1.3cm, right of=r,label={[rotate=90,xshift=-0.3cm,yshift=0.3cm]right:192*7*7}] {MBE6\_K5};
				\node (bp) [cell_r,yshift=-1.3cm,xshift=-1.3cm, right of=s,label={[rotate=90,xshift=0.1cm,yshift=0.3cm]right:320*7*7}] {Conv\_K1};
				\node (pool_fc) [cell_r,yshift=-1.3cm,xshift=-1.3cm, right of=bp,label={[rotate=90,xshift=-0.7cm,yshift=0.5cm]right:1280*7*7}] {Global Pooling + FC};
				\draw [arrow] (s1) -- (s2);
				\draw [arrow] (s2) -- (a);
				\draw [arrow] (a) -- (b);
				\draw [arrow] (b) -- (c);
				\draw [arrow] (c) -- (d);
				\draw [arrow] (d) -- (e);
				\draw [arrow] (e) -- (f);
				\draw [arrow] (f) -- (g);
				\draw [arrow] (g) -- (h);
				\draw [arrow] (h) -- (i);
				\draw [arrow] (i) -- (j);
				\draw [arrow] (j) -- (k);
				\draw [arrow] (k) -- (l);
				\draw [arrow] (l) -- (m);
				\draw [arrow] (m) -- (n);
				\draw [arrow] (n) -- (o);
				\draw [arrow] (o) -- (p);
				\draw [arrow] (p) -- (q);
				\draw [arrow] (q) -- (r);
				\draw [arrow] (r) -- (s);
				\draw [arrow] (s) --(bp);
				\draw [arrow] (bp) -- (pool_fc);
				\draw [ultra thick, xagray] (s1.south east)+(0.25,0.6) -- +(0.25,-2.2);
				\draw [ultra thick, xagray] (a.south east)+(0.25,0.5) --+(0.25,-2.3);
				\draw [ultra thick, xagray] (c.south east)+(0.25,0.5) --+(0.25,-2.3);
				\draw [ultra thick, xagray] (g.south east)+(0.25,0.5) --+(0.25,-2.3);
				\draw [ultra thick, xagray] (o.south east)+(0.25,0.2) --+(0.25,-2.6);
				\draw [dashed, xagray] (s2.south east)+(0.25,0.5) -- +(0.25,-2.3);
				\draw [dashed] (s.south east)+(0.25,0.2) -- +(0.25,-2.6);  
				\end{tikzpicture}
		}}
		\vskip -0.1in
		\subfigure{
			\centerline{
				\begin{tikzpicture}[thick,scale=0.5, every node/.style={scale=0.5},node distance=1.3cm]
				\node (s1) [cell_r,label={[rotate=90,xshift=0.1cm,yshift=0.5cm]right:3*224*224}] {Conv\_K3};
				\node (s2) [cell_r_b, yshift=-1.3cm,xshift=-0.3cm, xshift=-1cm, right of=s1,label={[rotate=90,xshift=0.1cm,yshift=0.3cm]right:32*112*112}] {MBE1\_K3};
				\node (a) [cell_r_g, e3, yshift=-1.3cm,xshift=-0.3cm, xshift=-1cm, right of=s2,label={[rotate=90,xshift=0.1cm,yshift=0.3cm]right:16*112*112}] {MBE3\_K5}; 
				\node (b) [cell_r_b,e3 , yshift=-1.3cm,xshift=-0.3cm, xshift=-1cm, right of=a,label={[rotate=90,xshift=0.1cm,yshift=0.3cm]right:32*56*56}] {MBE3\_K3};
				\node (c) [cell_r_g,e3 , yshift=-1.3cm,xshift=-0.3cm, xshift=-1cm, right of=b,label={[rotate=90,xshift=0.1cm,yshift=0.3cm]right:32*56*56}] {MBE3\_K5}; 
				\node (d) [cell_r_b,e3 , yshift=-1.3cm,xshift=-0.3cm, xshift=-1cm, right of=c,label={[rotate=90,xshift=0.1cm,yshift=0.3cm]right:40*28*28}] {MBE3\_K3};
				\node (e) [cell_r_b,e6 , yshift=-1.3cm,xshift=-0.3cm, xshift=-1cm, right of=d,label={[rotate=90,xshift=-0.3cm,yshift=0.5cm]right:40*28*28}] {MBE6\_K3};
				\node (f) [cell_r_g, e3 , yshift=-1.3cm,xshift=-0.3cm, xshift=-1cm, right of=e,label={[rotate=90,xshift=0.1cm,yshift=0.3cm]right:40*28*28}] {MBE3\_K5};
				\node (g) [cell_r_p, e3 , yshift=-1.3cm,xshift=-0.3cm, xshift=-1cm, right of=f,label={[rotate=90,xshift=0.1cm,yshift=0.3cm]right:40*28*28}] {MBE3\_K7};
				\node (h) [cell_r_b, e3 , yshift=-1.3cm,xshift=-0.3cm, xshift=-1cm, right of=g,label={[rotate=90,xshift=0.1cm,yshift=0.3cm]right:80*14*14}] {MBE3\_K3};
				\node (i) [cell_r_b, e6 , yshift=-1.3cm,xshift=-0.3cm, xshift=-1cm, right of=h,label={[rotate=90,xshift=-0.3cm,yshift=0.5cm]right:80*14*14}] {MBE6\_K3};
				\node (j) [cell_r_g, e3 , yshift=-1.3cm,xshift=-0.3cm, xshift=-1cm, right of=i,label={[rotate=90,xshift=0.1cm,yshift=0.3cm]right:80*14*14}] {MBE3\_K5};
				\node (k) [cell_r_b, e3 , yshift=-1.3cm,xshift=-0.3cm, xshift=-1cm, right of=j,label={[rotate=90,xshift=0.1cm,yshift=0.3cm]right:80*14*14}] {MBE3\_K3};
				\node (l) [cell_r_b, e6 , yshift=-1.3cm,xshift=-0.3cm, xshift=-1cm, right of=k,label={[rotate=90,xshift=-0.3cm,yshift=0.5cm]right:96*14*14}] {MBE6\_K3};
				\node (m) [cell_r_p, e3 , yshift=-1.3cm,xshift=-0.3cm, xshift=-1cm, right of=l,label={[rotate=90,xshift=0.1cm,yshift=0.5cm]right:96*14*14}] {MBE3\_K7}; 
				\node (n) [cell_r_b, e3 , yshift=-1.3cm,xshift=-0.3cm, xshift=-1cm, right of=m,label={[rotate=90,xshift=0.1cm,yshift=0.5cm]right:96*14*14}] {MBE3\_K3};
				\node (o) [cell_r_p, e6 , yshift=-1.3cm,xshift=-0.3cm, xshift=-1cm, right of=n,label={[rotate=90,xshift=-0.3cm,yshift=0.5cm]right:96*14*14}] {MBE6\_K7}; 
				\node (p) [cell_r_g, e6 , yshift=-1.3cm,xshift=-0.3cm, xshift=-1cm, right of=o,label={[rotate=90,xshift=-0.3cm,yshift=0.3cm]right:192*7*7}] {MBE6\_K5};
				\node (q) [cell_r_p, e6 , yshift=-1.3cm,xshift=-0.3cm, xshift=-1cm, right of=p,label={[rotate=90,xshift=-0.3cm,yshift=0.3cm]right:192*7*7}] {MBE6\_K7};
				\node (r) [cell_r_b, e6 , yshift=-1.3cm,xshift=-0.3cm, xshift=-1cm, right of=q,label={[rotate=90,xshift=-0.3cm,yshift=0.3cm]right:192*7*7}] {MBE6\_K3};
				\node (s) [cell_r_g, e6 , yshift=-1.3cm,xshift=-0.3cm, xshift=-1cm, right of=r,label={[rotate=90,xshift=-0.3cm,yshift=0.3cm]right:192*7*7}] {MBE6\_K5};
				\node (bp) [cell_r,yshift=-1.3cm,xshift=-0.3cm, xshift=-1cm, right of=s,label={[rotate=90,xshift=0.1cm,yshift=0.3cm]right:320*7*7}] {Conv\_K1};
				\node (pool_fc) [cell_r,yshift=-1.3cm,xshift=-0.3cm, xshift=-1cm, right of=bp,label={[rotate=90,xshift=-0.7cm,yshift=0.5cm]right:1280*7*7}] {Global Pooling + FC};
				\draw [arrow] (s1) -- (s2);
				\draw [arrow] (s2) -- (a);
				\draw [arrow] (a) -- (b);
				\draw [arrow] (b) -- (c);
				\draw [arrow] (c) -- (d);
				\draw [arrow] (d) -- (e);
				\draw [arrow] (e) -- (f);
				\draw [arrow] (f) -- (g);
				\draw [arrow] (g) -- (h);
				\draw [arrow] (h) -- (i);
				\draw [arrow] (i) -- (j);
				\draw [arrow] (j) -- (k);
				\draw [arrow] (k) -- (l);
				\draw [arrow] (l) -- (m);
				\draw [arrow] (m) -- (n);
				\draw [arrow] (n) -- (o);
				\draw [arrow] (o) -- (p);
				\draw [arrow] (p) -- (q);
				\draw [arrow] (q) -- (r);
				\draw [arrow] (r) -- (s);
				\draw [arrow] (s) --(bp);
				\draw [arrow] (bp) -- (pool_fc);
				\draw [ultra thick, xagray] (s1.south east)+(0.25,0.6) -- +(0.25,-2.2);
				\draw [ultra thick, xagray] (a.south east)+(0.25,0.5) --+(0.25,-2.3);
				\draw [ultra thick, xagray] (c.south east)+(0.25,0.5) --+(0.25,-2.3);
				\draw [ultra thick, xagray] (g.south east)+(0.25,0.5) --+(0.25,-2.3);
				\draw [ultra thick, xagray] (o.south east)+(0.25,0.2) --+(0.25,-2.6);
				\draw [dashed, xagray] (s2.south east)+(0.25,0.5) -- +(0.25,-2.3);
				\draw [dashed] (s.south east)+(0.25,0.2) -- +(0.25,-2.6);  
				\end{tikzpicture}
		}}
		\vskip -0.1in
		\subfigure{
			\centerline{
				\begin{tikzpicture}[thick,scale=0.5, every node/.style={scale=0.5},node distance=1.3cm]
				\node (s1) [cell_r,label={[rotate=90,xshift=0.1cm,yshift=0.5cm]right:3*224*224}] {Conv\_K3};
				\node (s2) [cell_r_b, yshift=-1.3cm,xshift=-0.3cm, xshift=-1cm, right of=s1,label={[rotate=90,xshift=0.1cm,yshift=0.3cm]right:32*112*112}] {MBE1\_K3};
				\node (a) [cell_r_g, e3, yshift=-1.3cm,xshift=-0.3cm, xshift=-1cm, right of=s2,label={[rotate=90,xshift=0.1cm,yshift=0.3cm]right:16*112*112}] {MBE3\_K5}; 
				\node (b) [cell_r_b,e3 , yshift=-1.3cm,xshift=-0.3cm, xshift=-1cm, right of=a,label={[rotate=90,xshift=0.1cm,yshift=0.3cm]right:32*56*56}] {MBE3\_K3};
				\node (c) [cell_r_p,e3 , yshift=-1.3cm,xshift=-0.3cm, xshift=-1cm, right of=b,label={[rotate=90,xshift=0.1cm,yshift=0.3cm]right:32*56*56}] {MBE3\_K7}; 
				\node (d) [cell_r_b,e3 , yshift=-1.3cm,xshift=-0.3cm, xshift=-1cm, right of=c,label={[rotate=90,xshift=0.1cm,yshift=0.3cm]right:40*28*28}] {MBE3\_K3};
				\node (e) [cell_r_b,e3 , yshift=-1.3cm,xshift=-0.3cm, xshift=-1cm, right of=d,label={[rotate=90,xshift=0.1cm,yshift=0.3cm]right:40*28*28}] {MBE3\_K3};
				\node (f) [cell_r_b, e3 , yshift=-1.3cm,xshift=-0.3cm, xshift=-1cm, right of=e,label={[rotate=90,xshift=0.1cm,yshift=0.3cm]right:40*28*28}] {MBE3\_K3};
				\node (g) [cell_r_b, e3 , yshift=-1.3cm,xshift=-0.3cm, xshift=-1cm, right of=f,label={[rotate=90,xshift=0.1cm,yshift=0.3cm]right:40*28*28}] {MBE3\_K3};
				\node (h) [cell_r_b, e3 , yshift=-1.3cm,xshift=-0.3cm, xshift=-1cm, right of=g,label={[rotate=90,xshift=0.1cm,yshift=0.3cm]right:80*14*14}] {MBE3\_K3};
				\node (i) [cell_r_b, e3 , yshift=-1.3cm,xshift=-0.3cm, xshift=-1cm, right of=h,label={[rotate=90,xshift=0.1cm,yshift=0.3cm]right:80*14*14}] {MBE3\_K3};
				\node (j) [cell_r_b, e6 , yshift=-1.3cm,xshift=-0.3cm, xshift=-1cm, right of=i,label={[rotate=90,xshift=-0.3cm,yshift=0.5cm]right:80*14*14}] {MBE6\_K3};
				\node (k) [cell_r_b, e3 , yshift=-1.3cm,xshift=-0.3cm, xshift=-1cm, right of=j,label={[rotate=90,xshift=0.1cm,yshift=0.3cm]right:80*14*14}] {MBE3\_K3};
				\node (l) [cell_r_b, e3 , yshift=-1.3cm,xshift=-0.3cm, xshift=-1cm, right of=k,label={[rotate=90,xshift=0.1cm,yshift=0.5cm]right:96*14*14}] {MBE3\_K3};
				\node (m) [cell_r_b, e3 , yshift=-1.3cm,xshift=-0.3cm, xshift=-1cm, right of=l,label={[rotate=90,xshift=0.1cm,yshift=0.5cm]right:96*14*14}] {MBE3\_K3}; 
				\node (n) [cell_r_b, e3 , yshift=-1.3cm,xshift=-0.3cm, xshift=-1cm, right of=m,label={[rotate=90,xshift=0.1cm,yshift=0.5cm]right:96*14*14}] {MBE3\_K3};
				\node (o) [cell_r_p, e6 , yshift=-1.3cm,xshift=-0.3cm, xshift=-1cm, right of=n,label={[rotate=90,xshift=-0.3cm,yshift=0.5cm]right:96*14*14}] {MBE6\_K7}; 
				\node (p) [cell_r_p, e6 , yshift=-1.3cm,xshift=-0.3cm, xshift=-1cm, right of=o,label={[rotate=90,xshift=-0.3cm,yshift=0.3cm]right:192*7*7}] {MBE6\_K7};
				\node (q) [cell_r_b, e6 , yshift=-1.3cm,xshift=-0.3cm, xshift=-1cm, right of=p,label={[rotate=90,xshift=-0.3cm,yshift=0.3cm]right:192*7*7}] {MBE6\_K3};
				\node (r) [cell_r_b, e6 , yshift=-1.3cm,xshift=-0.3cm, xshift=-1cm, right of=q,label={[rotate=90,xshift=-0.3cm,yshift=0.3cm]right:192*7*7}] {MBE6\_K3};
				\node (s) [cell_r_g, e6 , yshift=-1.3cm,xshift=-0.3cm, xshift=-1cm, right of=r,label={[rotate=90,xshift=-0.3cm,yshift=0.3cm]right:192*7*7}] {MBE6\_K5};
				\node (bp) [cell_r,yshift=-1.3cm,xshift=-0.3cm, xshift=-1cm, right of=s,label={[rotate=90,xshift=0.1cm,yshift=0.3cm]right:320*7*7}] {Conv\_K1};
				\node (pool_fc) [cell_r,yshift=-1.3cm,xshift=-0.3cm, xshift=-1cm, right of=bp,label={[rotate=90,xshift=-0.7cm,yshift=0.5cm]right:1280*7*7}] {Global Pooling + FC};
				\draw [arrow] (s1) -- (s2);
				\draw [arrow] (s2) -- (a);
				\draw [arrow] (a) -- (b);
				\draw [arrow] (b) -- (c);
				\draw [arrow] (c) -- (d);
				\draw [arrow] (d) -- (e);
				\draw [arrow] (e) -- (f);
				\draw [arrow] (f) -- (g);
				\draw [arrow] (g) -- (h);
				\draw [arrow] (h) -- (i);
				\draw [arrow] (i) -- (j);
				\draw [arrow] (j) -- (k);
				\draw [arrow] (k) -- (l);
				\draw [arrow] (l) -- (m);
				\draw [arrow] (m) -- (n);
				\draw [arrow] (n) -- (o);
				\draw [arrow] (o) -- (p);
				\draw [arrow] (p) -- (q);
				\draw [arrow] (q) -- (r);
				\draw [arrow] (r) -- (s);
				\draw [arrow] (s) --(bp);
				\draw [arrow] (bp) -- (pool_fc);
				\draw [ultra thick, xagray] (s1.south east)+(0.25,0.6) -- +(0.25,-2.2);
				\draw [ultra thick, xagray] (a.south east)+(0.25,0.5) --+(0.25,-2.3);
				\draw [ultra thick, xagray] (c.south east)+(0.25,0.5) --+(0.25,-2.3);
				\draw [ultra thick, xagray] (g.south east)+(0.25,0.5) --+(0.25,-2.3);
				\draw [ultra thick, xagray] (o.south east)+(0.25,0.2) --+(0.25,-2.6);
				\draw [dashed, xagray] (s2.south east)+(0.25,0.5) -- +(0.25,-2.3);
				\draw [dashed] (s.south east)+(0.25,0.2) -- +(0.25,-2.6);  
				\end{tikzpicture}
		}}
		\caption{Architectures of FairNAS-A,B,C (from top to bottom). MBE$x$\_K$y$ means an expansion rate of $x$ and a kernel size of $y$ for its depthwise convolution} 
		\label{fig:fairnas-a}
	\end{center}
\end{figure*}
%
%

\subsection{Hyperparameters for MoreMNAS variant}
We list the hyperparameters for the adopted MoreMNAS \cite{chu2019multi} variant in Table \ref{tab:pipelienhyper}. It has a population N of 64 models. It also takes a hierarchical mutation strategy. Respectively, $p_{rm}, p_{re}, p_{pr}$ indicate probabilities for random mutation, reinforce mutation and prior regulator, where $p_{re}$ again is divided into $p_{K-M}$ for \emph{roulette wheel selection}, and $p_{M}$  for reinforced controller.

\begin{table}[ht]
	\begin{center}
		\begin{small}
				\begin{tabular}{lclc}
					\toprule
					Item & value & Item & value \\
					\midrule
					Population N & 64 & Mutation Ratio  & 0.8 \\
					$p_{rm}$ & 0.2 & $p_{re}$   & 0.65 \\
					$p_{pr}$ & 0.15 & $p_{M}$ & 0.7 \\
					$p_{K-M}$ & 0.3 \\
					\bottomrule
				\end{tabular}
		\end{small}
	\end{center}
	\caption{Hyperparameters for the second-stage EA search.}
	\label{tab:pipelienhyper}
\end{table}

\subsection{Training of stand-alone models}

\begin{figure}[ht]
\centering
\subfigure{
\begin{tikzpicture}[spy using outlines={circle,xagray,magnification=3,size=1.6cm, connect spies}]
\node {
\includegraphics[scale=0.6]{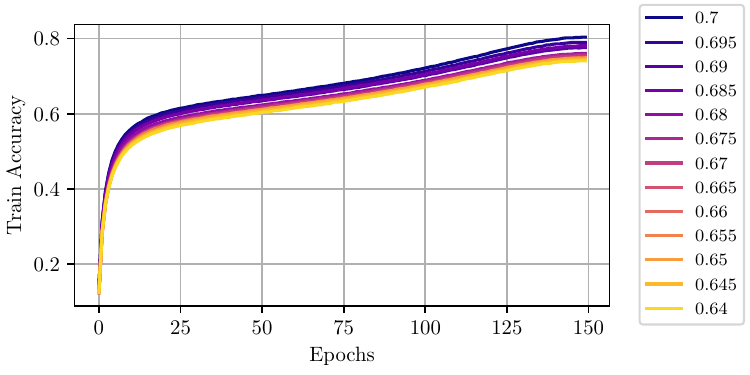}};
\spy on (2.13,1.38) in node [left] at (1.7,0.2);
\end{tikzpicture}
}
\vskip -0.2in
\subfigure{
\begin{tikzpicture}[spy using outlines={circle,xagray,magnification=5,size=1.5cm, connect spies}]
\node {\includegraphics[scale=0.6]{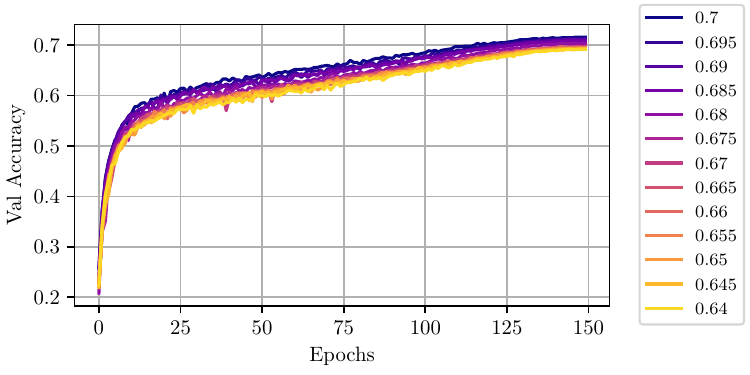}};
\spy on (2.15,1.46) in node [left] at (1.6,0.3);
\end{tikzpicture}
}
\caption{Train and validation accuracies (ground truth) of all 13 stand-alone models when being fully trained with the same hyperparameters. Lines are labelled with corresponding one-shot accuracies (predicted) sorted in descending order (as reflected by color gradient).}
\label{fig:standalone_val}
\end{figure}

We picked 13 models to train from scratch whose one-shot accuracies are approximately evenly spaced, ranging in $[0.641, 0.7]$. We keep the exactly same hyperparameters as the supernet training. Their corresponding stand-alone accuracies are within $[0.692, 0.715]$. Figure~\ref{fig:standalone_val} plots the training process, from which we observe the ranking of one-shot models are generally maintained. 
The model-meta (indices of operations) of these 13 models are listed in Table~\ref{tab:13models}. Besides, the mapping from an index in model-meta to a searchable operation is given in Table~\ref{tab:meta-mapping}.

\setlength{\tabcolsep}{4pt}
\begin{table}[ht]
	\begin{center}
		\begin{small}
				\begin{tabular}{lc}
					\toprule
					Index & Model Meta \\
					\midrule
					0 & [0, 0, 0, 0, 0, 0, 0, 0, 0, 0, 0, 0, 0, 0, 0, 0] \\
					1 & [0, 0, 0, 0, 0, 0, 0, 0, 0, 0, 0, 0, 0, 1, 0, 0] \\
					2 & [0, 0, 0, 0, 0, 0, 0, 1, 0, 0, 0, 0, 0, 1, 0, 0] \\
					3 & [0, 0, 0, 0, 1, 0, 1, 0, 0, 0, 1, 0, 0, 1, 0, 0] \\
					4 & [0, 0, 0, 0, 0, 0, 1, 1, 0, 1, 1, 0, 0, 1, 0, 0] \\
					5 & [0, 1, 0, 1, 0, 0, 1, 1, 0, 0, 0, 0, 0, 1, 0, 0] \\
					6 & [0, 1, 0, 0, 1, 1, 1, 1, 1, 1, 0, 0, 0, 1, 2, 0] \\
					7 & [0, 1, 0, 1, 1, 4, 1, 1, 1, 1, 1, 0, 0, 1, 2, 0] \\
					8 & [0, 1, 4, 1, 0, 3, 1, 1, 1, 1, 1, 1, 0, 2, 0, 0] \\
					9 & [0, 1, 0, 0, 1, 5, 1, 1, 0, 5, 1, 1, 0, 1, 2, 3] \\
					10 &[3, 1, 4, 1, 3, 4, 1, 4, 1, 3, 1, 1, 3, 1, 2, 0]\\
					11 & [0, 1, 4, 3, 1, 3, 1, 1, 1, 3, 4, 1, 3, 1, 2, 3]\\
					12 & [1, 5, 3, 2, 1, 4, 3, 4, 1, 5, 1, 1, 3, 5, 5, 3]\\
					\bottomrule
				\end{tabular}
		\end{small}
	\end{center}
	\caption{Model-meta of  13 sampled stand-alone models for ranking analysis.}
	\label{tab:13models}
\end{table}

\begin{table}[ht]
	\begin{center}
		\begin{small}
				\begin{tabular}{lcc}
					\toprule
					Model Meta Index & kernel & Expansion Rate \\
					\midrule
					0 & 3&3 \\
					1 & 5&3 \\
					2 & 7&3 \\
					3 & 3&6 \\
					4 & 5&6 \\
					5 & 7&6 \\
					\bottomrule
				\end{tabular}
		\end{small}
	\end{center}
	\caption{Mapping between model-meta index and operations}
	\label{tab:meta-mapping}
\end{table}

\subsection{Evolutionary Searching}
\begin{figure}[ht]
\centering
\subfigure{}{
	\centering
	\includegraphics[scale=0.6]{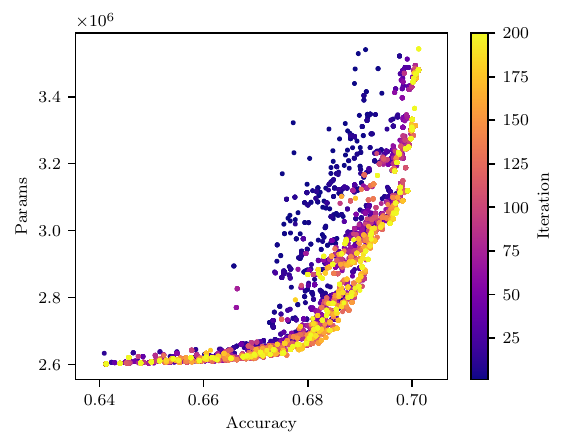}
	\label{fig:params-acc}
}
\subfigure{}{
	\centering
	\includegraphics[scale=0.6]{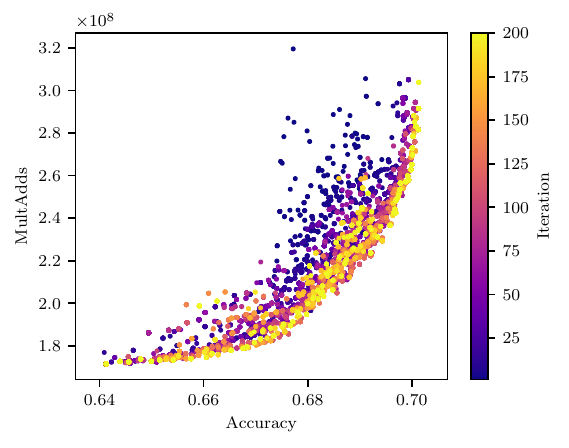}
	}
\caption{FairNAS evolution process of 200 generations, with 64 models sampled in each generation. Number of parameters, multiply-adds are charted with top-1 accuracies on the ImageNet validation set. }
\label{fig:fairnas-evolution}
\end{figure}

The evolutionary search of FairNAS based on MoreMNAS variant \cite{chu2019multi} is shown in Figure~\ref{fig:fairnas-evolution}. At each generation, 64 models are evaluated by our fair supernet, after 200 generations, the evolution converges, the Pareto-front is shown in bright yellow, each dot represents a candidate network.


\subsection{Object Detection}

For object detection, we treat FairNAS models as drop-in replacements for RetinaNet's backbone \cite{lin2017focal}. We follow the same setting as \cite{lin2017focal} and exploit MMDetection toolbox \cite{chen2019mmdetection} for training. All the models are trained and evaluated on MS COCO dataset (train2017 and val2017 respectively) \cite{lin2014coco} for 12 epochs with a batch size of 16. The initial learning rate is 0.01 and decayed by 0.1$\times$ at epochs 8 and 11.   


All baselines in the paper are mobile networks. The input features from these backbones to the FPN module  are from the last depthwise layers of stage 2 to 5\footnote{We follow the typical nomination for the definition of stages and the orders start from 1.}.  The number of  output channels of FPN is kept 256 as \cite{lin2017focal}. We also use $\alpha=0.25$ and $\gamma=2.0$ for the focal loss. Given longer training epochs and other tricks, the detection performance can be improved further. However, it's sufficient to compare the transferability of various methods.

\end{document}